\documentclass{article}
\usepackage{spconf,amsmath,graphicx,hyperref}

\usepackage[table]{xcolor}
\usepackage{graphicx}
\usepackage{subcaption}
\usepackage{indentfirst}
\usepackage{amssymb} 

\title{SpArSe-Up: Learnable Sparse Upsampling for 3D Generation with High-Fidelity Textures}
\name{Lu Xiao\textsuperscript{$\dagger$}, Jiale Zhang\textsuperscript{$\dagger$}, Yang Liu, Taicheng Huang\textsuperscript{*}, Xin Tian \thanks{\textsuperscript{$\dagger$}Equal contribution . \textsuperscript{*}Corresponding author: huangtaicheng.1@jd.com }}
\address{JD.com, Beijing, China}
\begin{document}
%
\maketitle
\begin{abstract}
The creation of high-fidelity 3D assets is often hindered by a `pixel-level pain point': the loss of high-frequency details. Existing methods often trade off one aspect for another: either sacrificing cross-view consistency, resulting in torn or drifting textures, or remaining trapped by the resolution ceiling of explicit voxels, forfeiting fine texture detail. In this work, we propose \textit{Sparse-Up}, a memory-efficient, high-fidelity texture modeling framework that effectively preserves high-frequency details. We use sparse voxels to guide texture reconstruction and ensure multi-view consistency, while leveraging \textit{surface anchoring} and \textit{view-domain partitioning} to break through resolution constraints. \textit{Surface anchoring} employs a learnable upsampling strategy to constrain voxels to the mesh surface, eliminating over 70\% of redundant voxels present in traditional voxel upsampling. \textit{View-domain partitioning} introduces an image patch-guided voxel partitioning scheme, supervising and back-propagating gradients only on visible local patches. Through these two strategies, we can significantly reduce memory consumption during high-resolution voxel training without sacrificing geometric consistency, while preserving high-frequency details in textures.
\end{abstract}
\section{Introduction}
\label{sec:intro}
3D assets are widely used in the fields of games, films, and e-commerce. In recent years, generative artificial intelligence has achieved remarkable progress in areas such as text\cite{achiam2023gpt,touvron2023llama}, images\cite{rombach2022high,ho2020denoising}, videos\cite{wan2025wan,kong2024hunyuanvideo}, and 3D content\cite{kerbl20233d,zhang2024clay,hunyuan3d2025hunyuan3d,xiang2025structured,li2025step1x}. Due to the inherent complexity of 3D geometric representation, the generation of 3D content is even more challenging.



Current 3D generation models primarily follow two architectures: two-stage methods compress 3D shapes into latent token sequences for geometry and use multi-view diffusion for texture, but lack explicit geometric constraints, leading to misaligned and blurry textures. Sparse voxel-based methods encode local features via active voxels but face a trade-off: low resolution misses high-frequency details, while high resolution demands prohibitive memory, limiting detail quality.

Building on the sparse voxel generation framework\cite{xiang2025structured}, we seek to break voxel resolution limits and achieve high-fidelity texture modeling. We note that traditional voxel-upscaling algorithms usually start from original voxel vertices and generate eight new ones, but this creates many off-surface points, leading to significant redundancy in voxel representation. By statistically analyzing the training data, we find that 70\% of these points are off-surface; such redundancies become the main computational and memory bottleneck in high-resolution training, severely limiting further increases in voxel resolution.


To address this issue, we propose a learnable upsampling scheme called \textit{surface anchoring}. By creating a learnable mapping between redundant upsampled points and true on-surface locations, and designing a network architecture symmetric to the Texture-VAE decoder for end-to-end learning, we effectively eliminate redundant points and boost the efficiency and fidelity of voxel representation. 

Nevertheless, high resolutions cause voxel grid points to explode and 3D Gaussians from the Texture-VAE decoder to grow accordingly, keeping GPU memory consumption during rendering high. To further reduce computational overhead, we propose a \textit{view-domain partitioning} strategy: using image patches as units, it culls voxels so only those within the current patch participate in rendering and back-propagation. This approach, leveraging localized rendering and gradient updates, significantly lowers memory usage. 

\begin{figure*}[t]
    \centering
    \includegraphics[width=\linewidth]{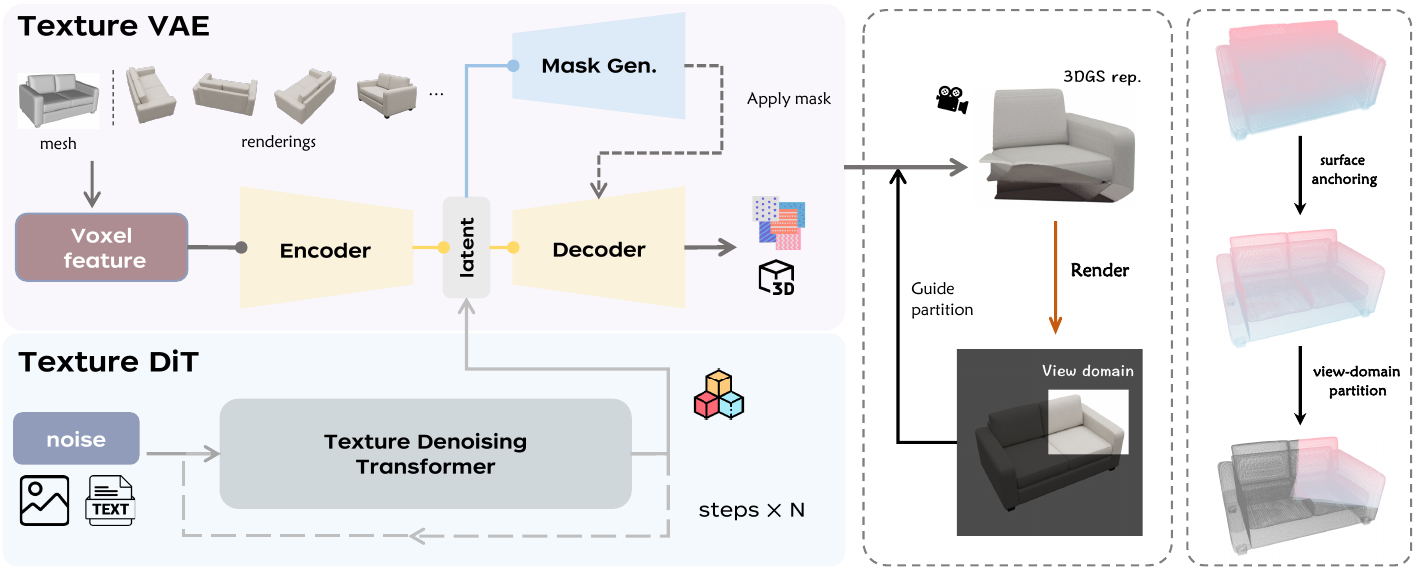}
    \caption{\textbf{Overall workflow of the Sparse-Up 3D generation architecture.} The workflow modules include Texture-VAE, Mask Generator and Texture-DiT. The right side illustrates the process of removing redundant points to achieve surface anchoring using a MASK generator, as well as local rendering and supervision performed via view-domain partitioning.}
    \label{fig:pipeline}
    \vspace{-1em}
    \label{pipeline}
\end{figure*}

In summary, our main contributions are:
\begin{itemize}
    \item We propose a \textit{surface anchoring} scheme that learns a mapping from redundant upsampled points to true surface points, eliminating redundancy and enhancing voxel efficiency.  
    \item We introduce a \textit{view-domain partitioning} strategy that performs localized rendering and reconstruction, significantly lowering memory consumption during rendering.
    \item Experiments demonstrate that our approach has attained state-of-the-art performance in the field of texture reconstruction.
\end{itemize}

\section{METHODOLOGY}
\label{sec:method}

We introduce Sparse-Up, an end-to-end framework that supports high-resolution and high-fidelity texture modeling, with the overall pipeline illustrated in fig.\ref{pipeline}. The framework takes sparse voxels embedded with texture features as input and, after a VAE encoder–decoder pass, directly outputs an explicit 3D Gaussian representation (3DGS). This explicit geometry inherently guarantees multi-view consistency. To break the resolution bottleneck, we propose two complementary schemes: \textit{surface anchoring} and \textit{view-domain partitioning}. Section \ref{sec:Surface-Anchoring Upsampling} presents the learnable surface-anchoring upsampling strategy that accurately maps redundant voxels onto the object surface, while Section \ref{sec:View-domain Partitioning} details the view-domain partitioning mechanism, which performs localized rendering and gradient back-propagation on a per-patch basis to reduce memory consumption.

\subsection{Surface-Anchoring Upsampling}
\label{sec:Surface-Anchoring Upsampling}

To achieve high-resolution reconstruction, we integrate two upsampling modules into the decoder architecture. Traditional upsampling often produces numerous off-surface and redundant points, degrading reconstructed surface quality and causing high memory consumption. To tackle these issues, we introduce an advanced transformer-based Mask Generator to guide redundancy elimination.

\begin{figure}[t]
\centering
\subfloat[]{
		\includegraphics[scale=0.12]{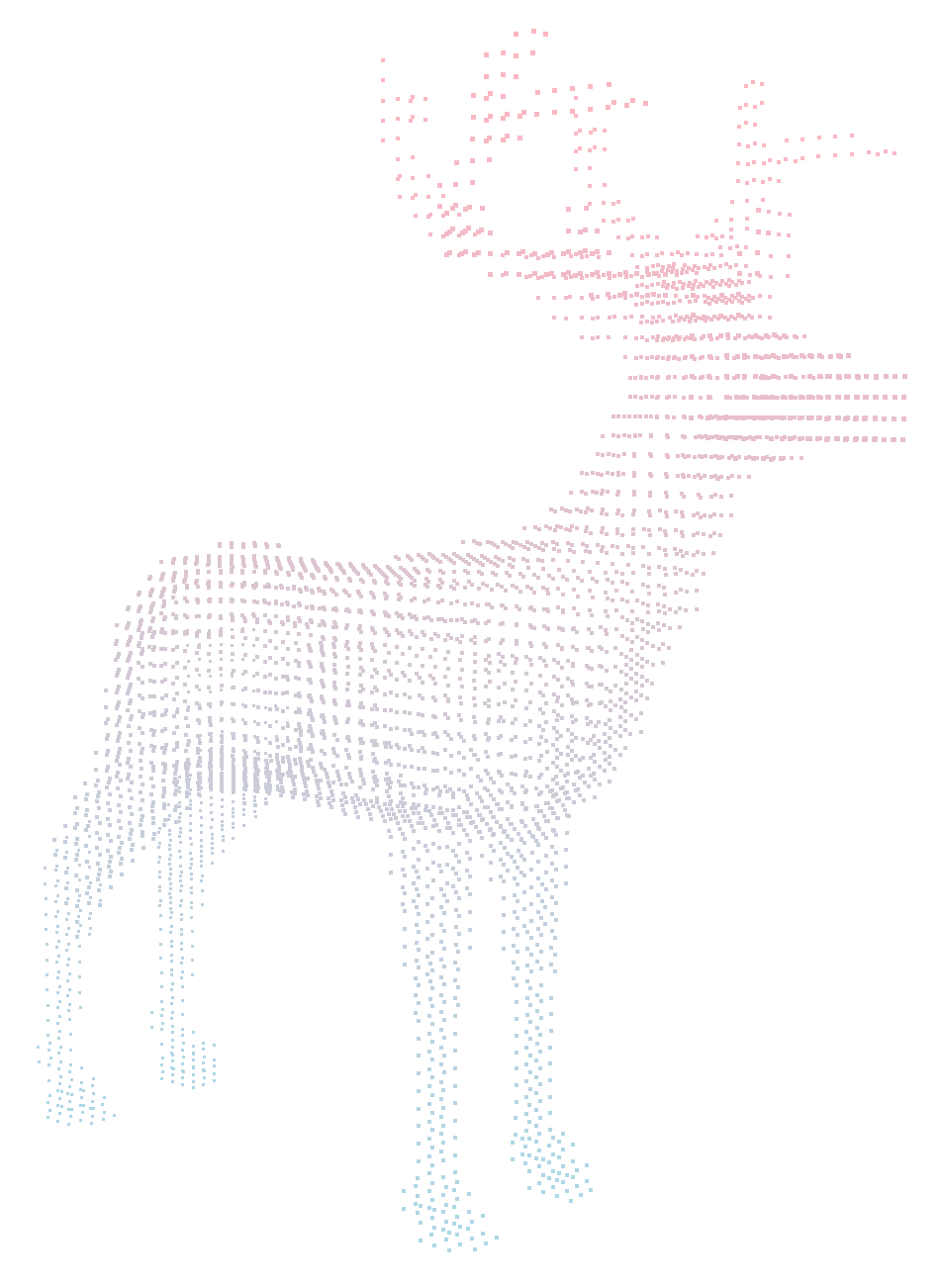}}\hspace{2em}
\subfloat[]{
		\includegraphics[scale=0.12]{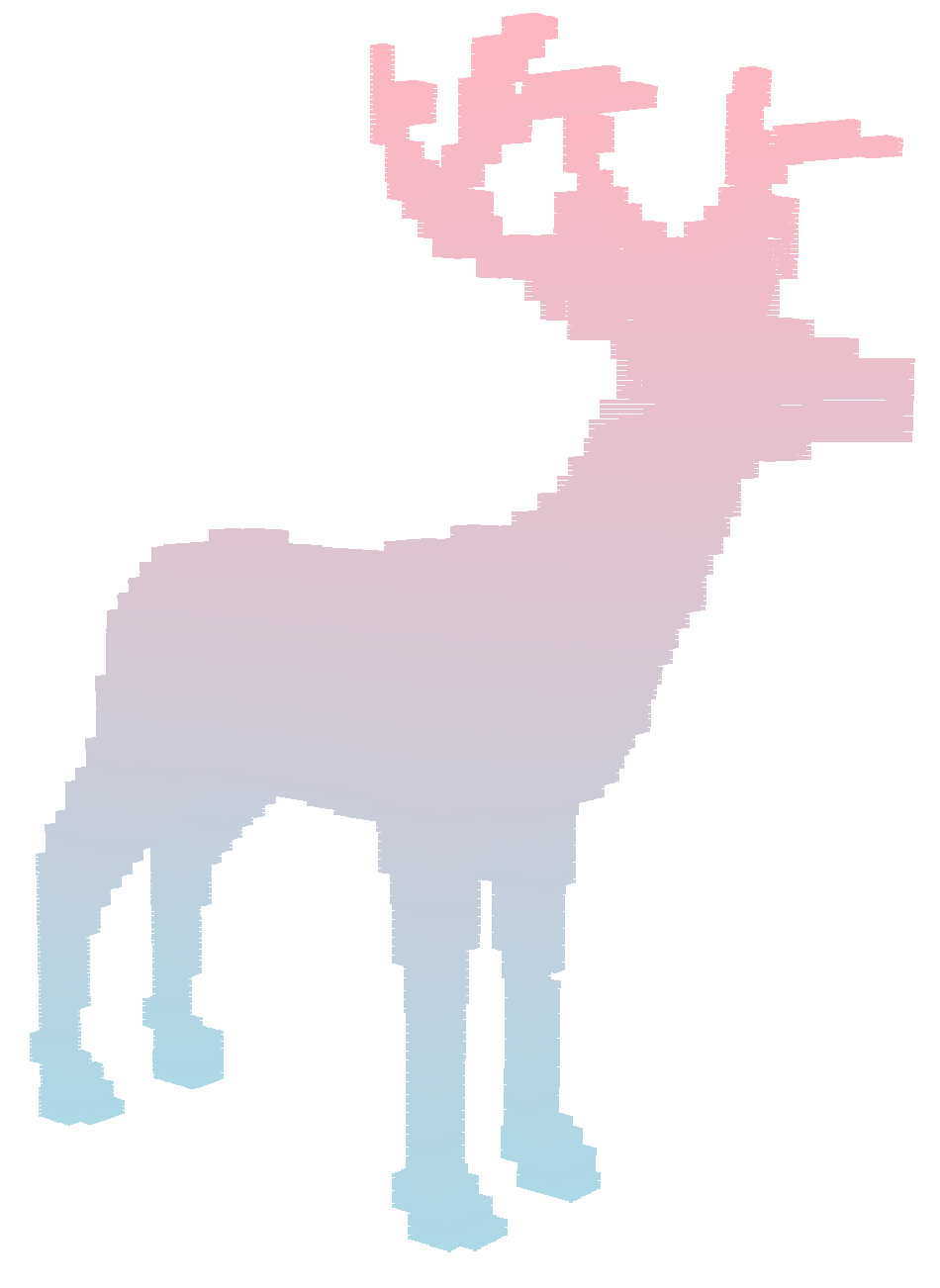}}\hspace{2em}
\subfloat[]{
		\includegraphics[scale=0.12]{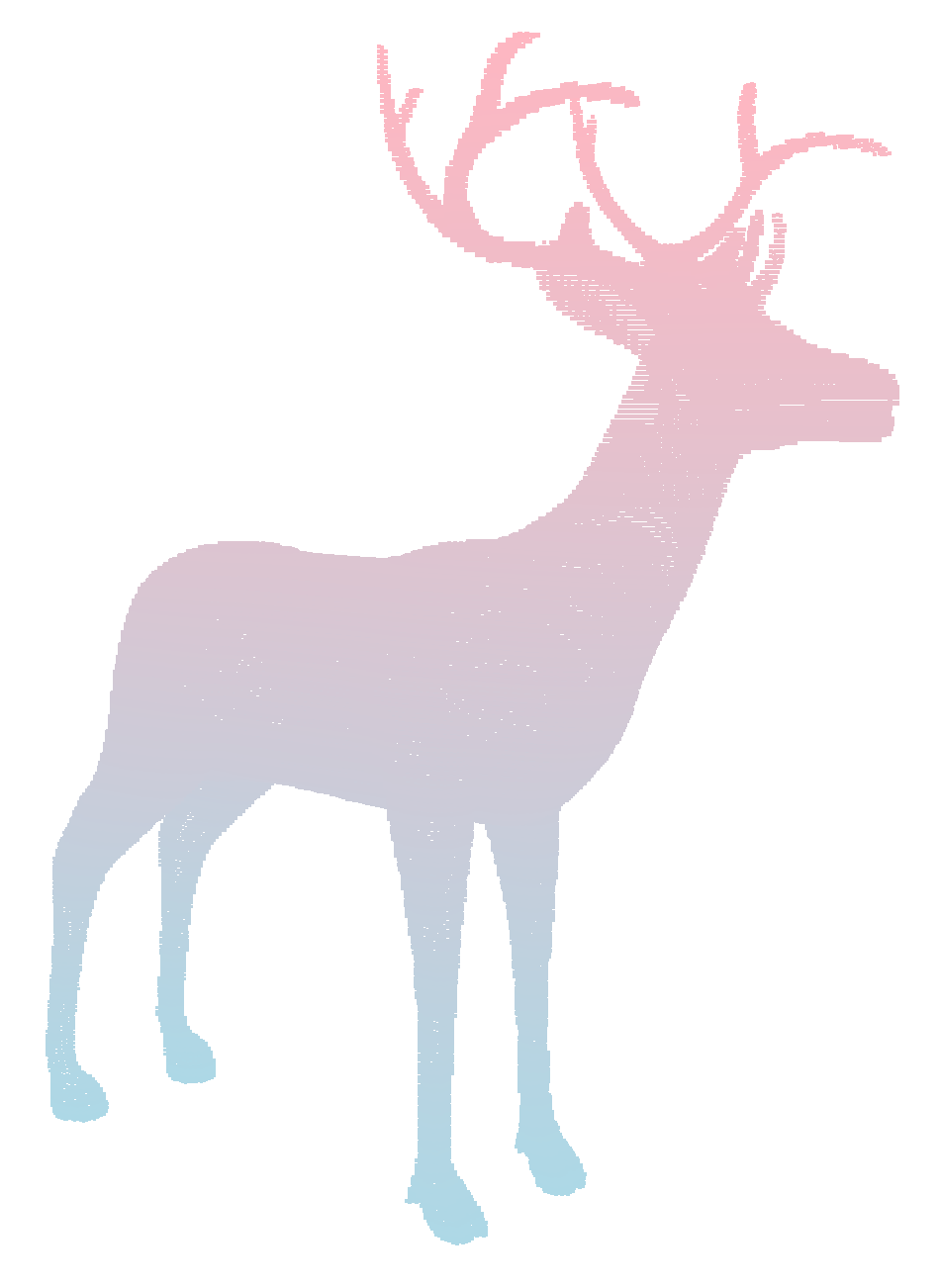}}
\caption{\textbf{Comparison of voxel visualizations.} (a) Low-resolution original data; (b) Traditional upsampling result (with redundancy); (c) High-resolution data after redundancy removal via generated mask, with approximately 70\% of points reduced from (b). It can be observed that high-resolution data contains much richer information than low-resolution data, and the generated mask has effectively eliminated a large number of redundant voxels.}

\vspace{-1.2 em}
\label{fig: voxel visualization.}
\end{figure}

Specifically, this component explicitly predicts a mapping mask to transform redundant points into their on-surface counterparts in each upsampling module. This mapping then guides the 3DGS decoder for precise upsampling—removing redundancy while preserving fine surface detail integrity (as shown in fig.\ref{fig: voxel visualization.}). Thus, our method boosts reconstructed surface spatial resolution, retains intricate textures and geometric features critical for high-fidelity 3D models, and reduces memory consumption by eliminating nearly 70\% of voxels from traditional voxel upsampling, enabling higher-resolution training.

To explicitly supervise the mask generator, we create ground-truth (GT) masks of different resolutions for 3D models in our dataset. Here, we illustrate the process using a $128$-resolution mask as an example. First, each model is voxelized at 64 and 128 resolutions to get surface voxel sets $V_{64}$ and $V_{128} $. Next, $V_{64}$ undergo a traditional upsampling process to produce $\hat{V}_{128}$ — a 128-resolution voxel set with redundancies, which fully contains $V_{128}$. Finally, by performing an intersection operation on the voxel coordinates of ${V}_{128}$ and $\hat{V}_{128}$, we can derive the GT mask ${M}_{128}$ that maps $\hat{V}_{128}$ to ${V}_{128}$, as illustrated by the following equation:
\begin{equation}
{M}_{128}(v) = 
\begin{cases} 
1, & \text{if } v \in V_{128} \\ 
0, & \text{otherwise}
\end{cases}
,\quad \forall v \in \hat{V}_{128}\,,
\end{equation}
Note that $M_{128}$ has the same shape as $\hat{V}_{128}$. Through this process, we generate multi-resolution GT masks for each 3D model to supervise the mask generator. However, the voxel arrangement in $\hat{M}_{128}$ generated by the mask generator via sparse convolution may differ from $M_{128}$. To ensure precise supervision, we apply a hash mapping from $M_{128}$ to $\hat{M}_{128}$ to align their order for accurate supervision.  

This approach enables precise supervision of the mask generator; by leveraging multi-resolution GT masks, the generator is better able to predict masks that remove redundancy while preserving key surface details.


\begin{figure}[t]
\captionsetup[subfloat]{labelsep=none,format=plain,labelformat=empty}
\centering
\subfloat{
    \includegraphics[width=0.11\textwidth]{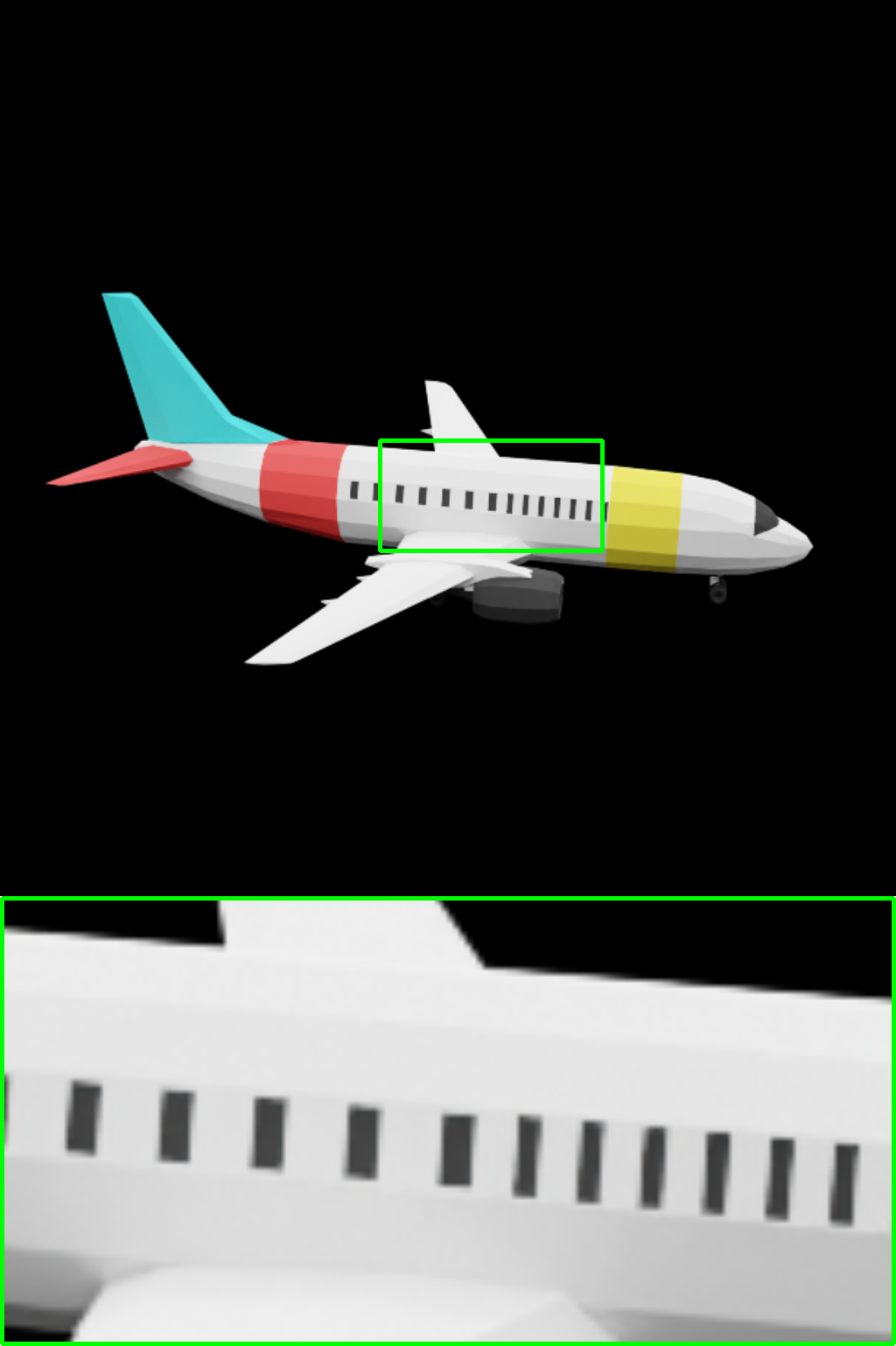} 
}
\subfloat{
    \includegraphics[width=0.11\textwidth]{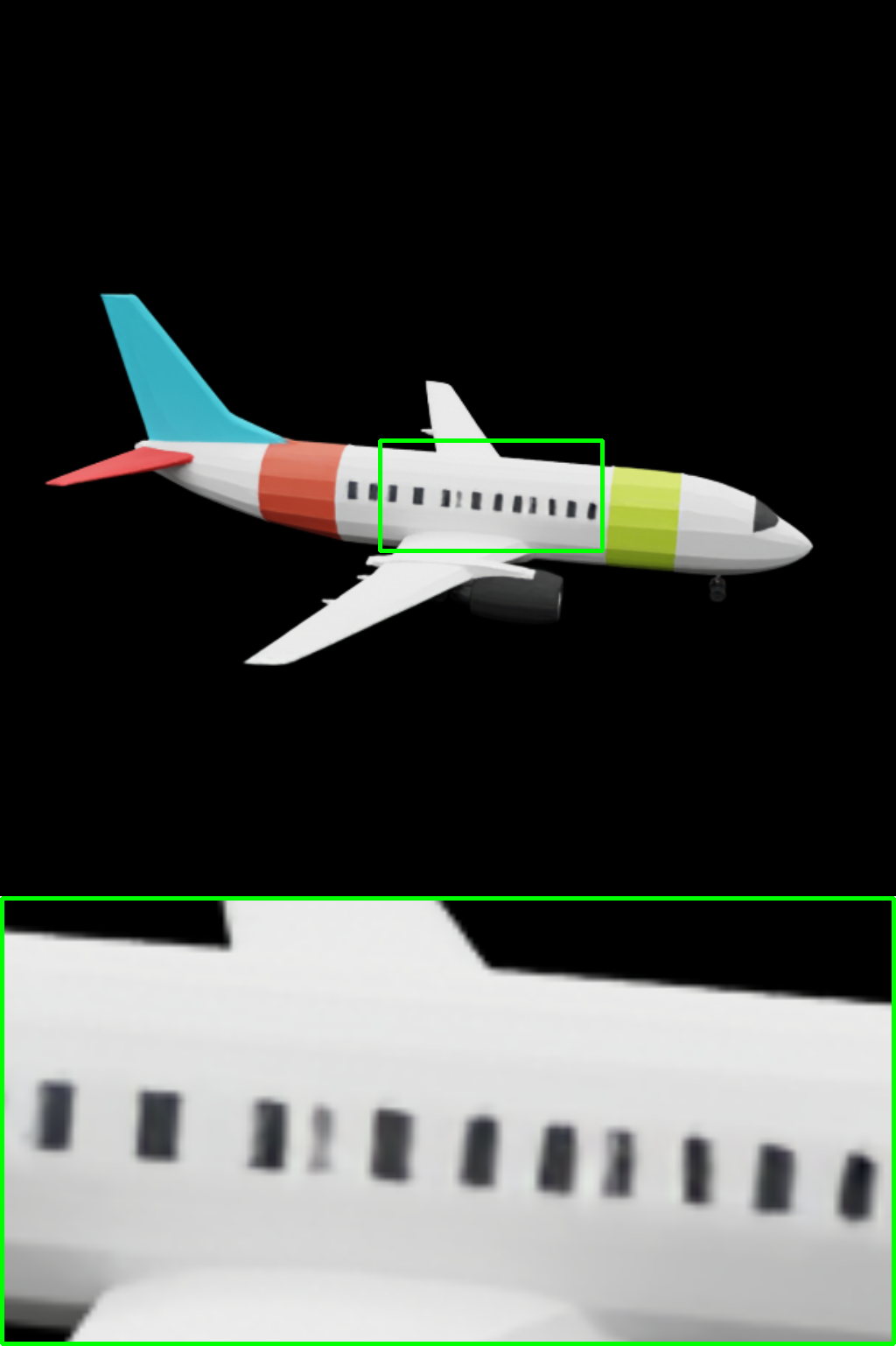}
}
\subfloat{
    \includegraphics[width=0.11\textwidth]{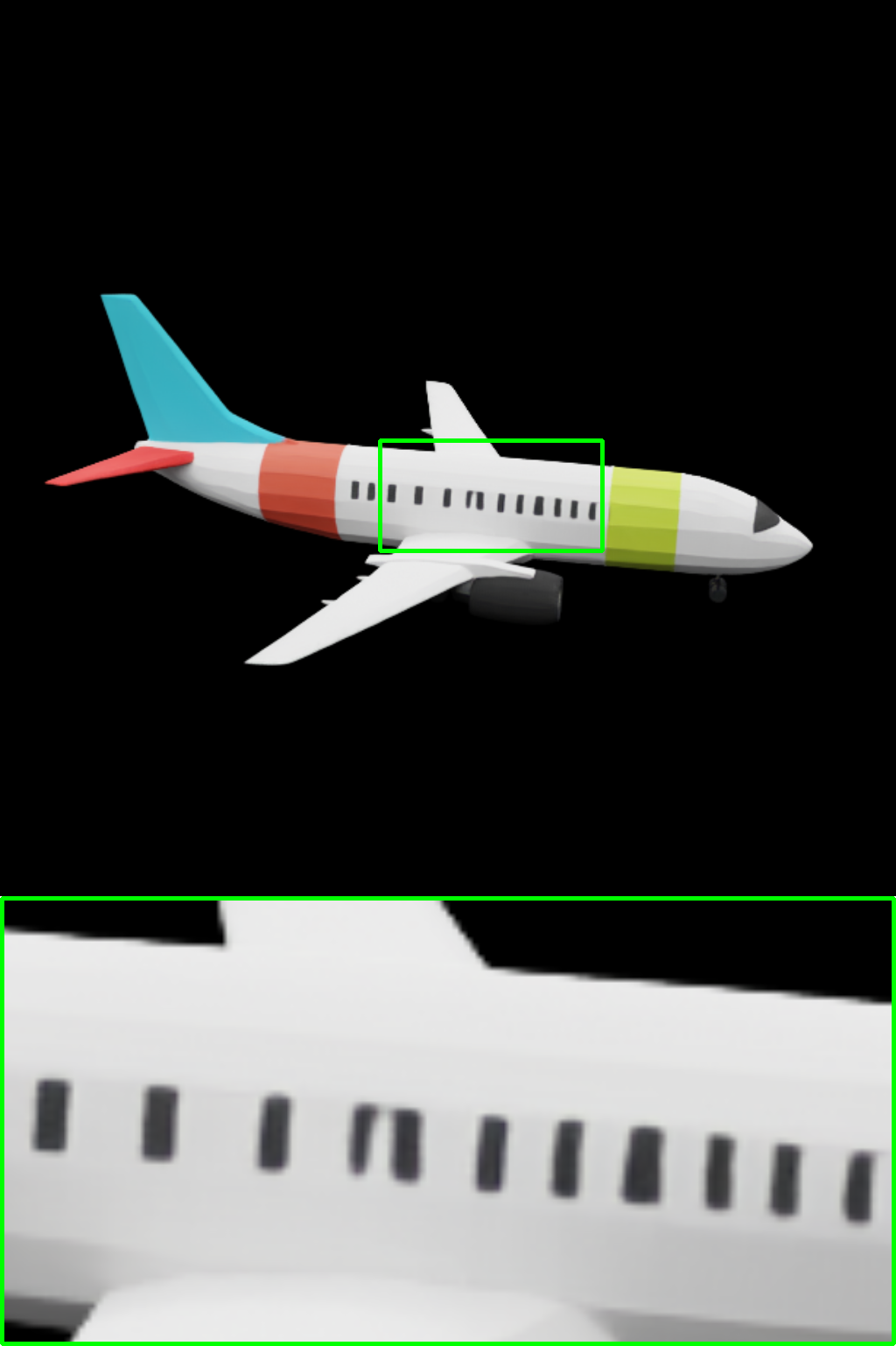}
}
\subfloat{
    \includegraphics[width=0.11\textwidth]{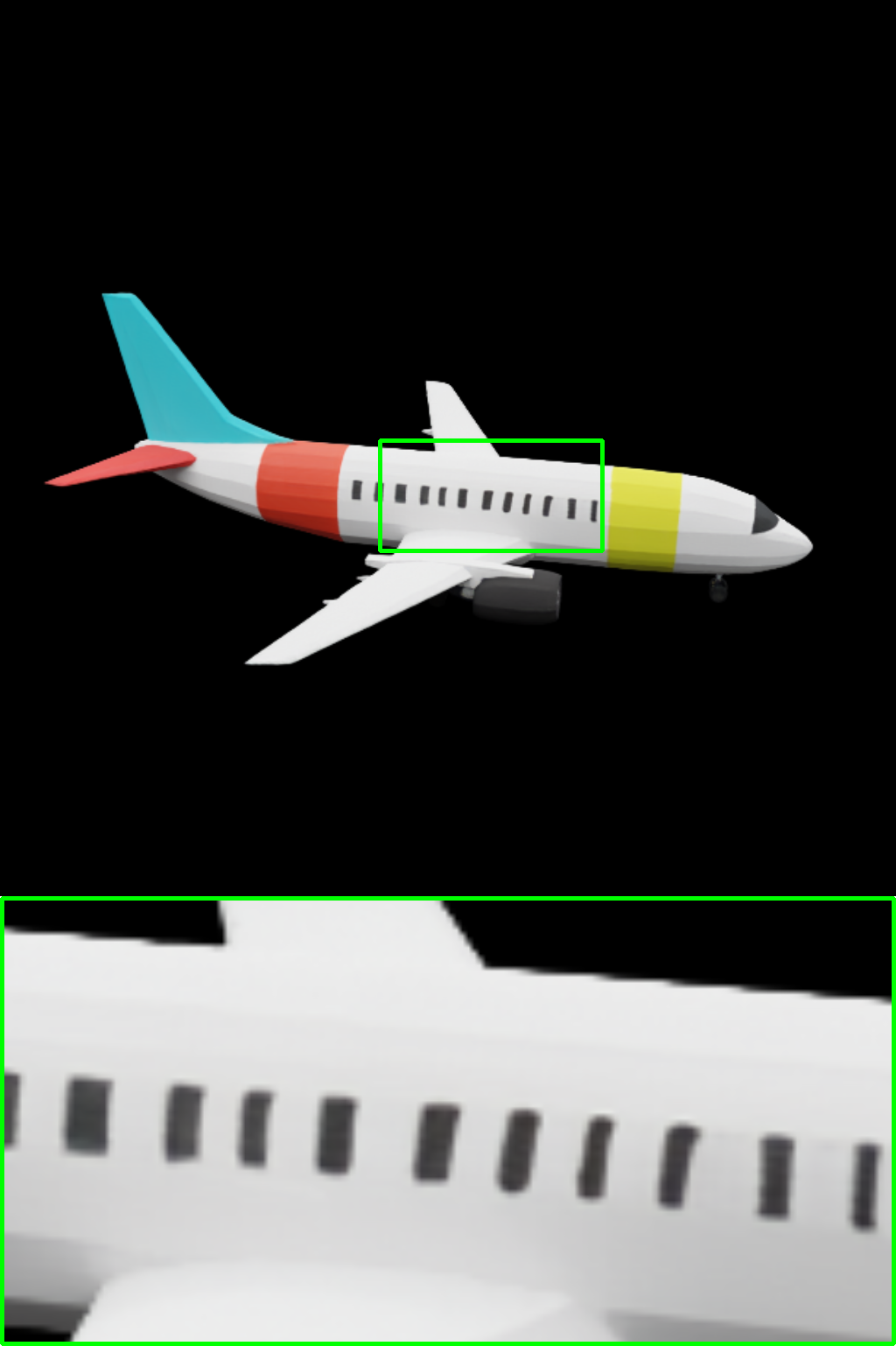}
}

\subfloat{
    \includegraphics[width=0.11\textwidth]{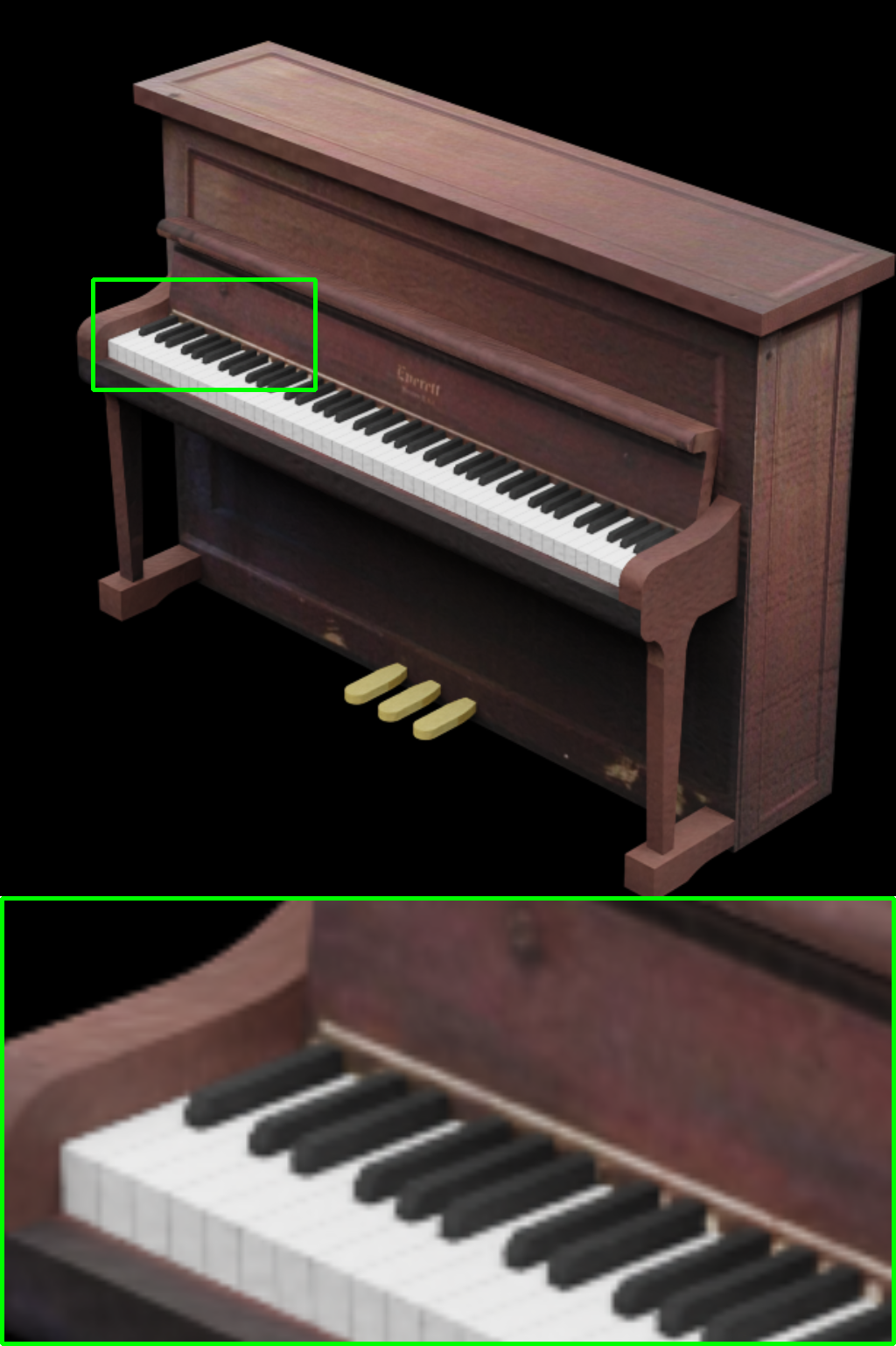} 
}
\subfloat{
    \includegraphics[width=0.11\textwidth]{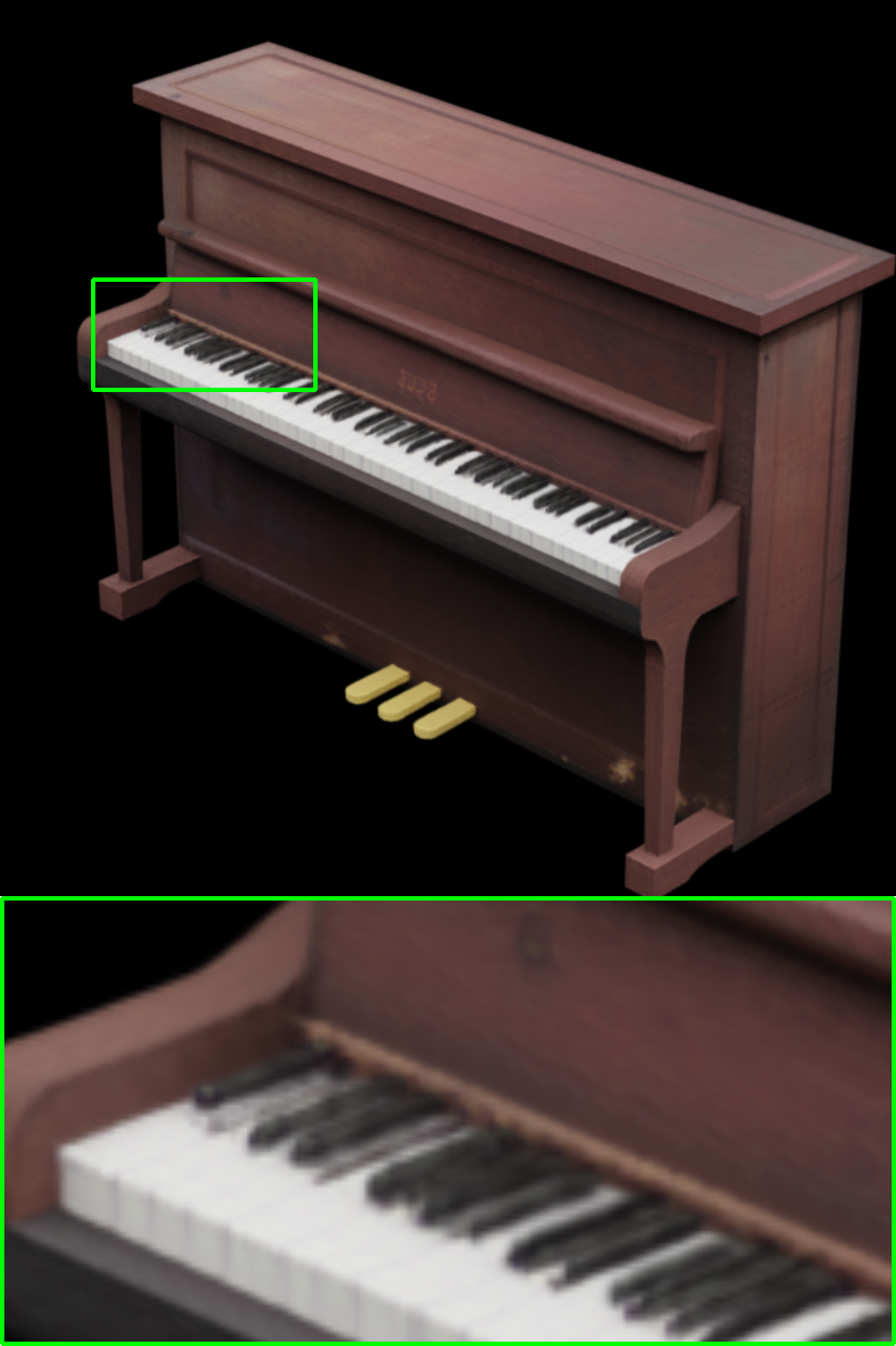}
}
\subfloat{
    \includegraphics[width=0.11\textwidth]{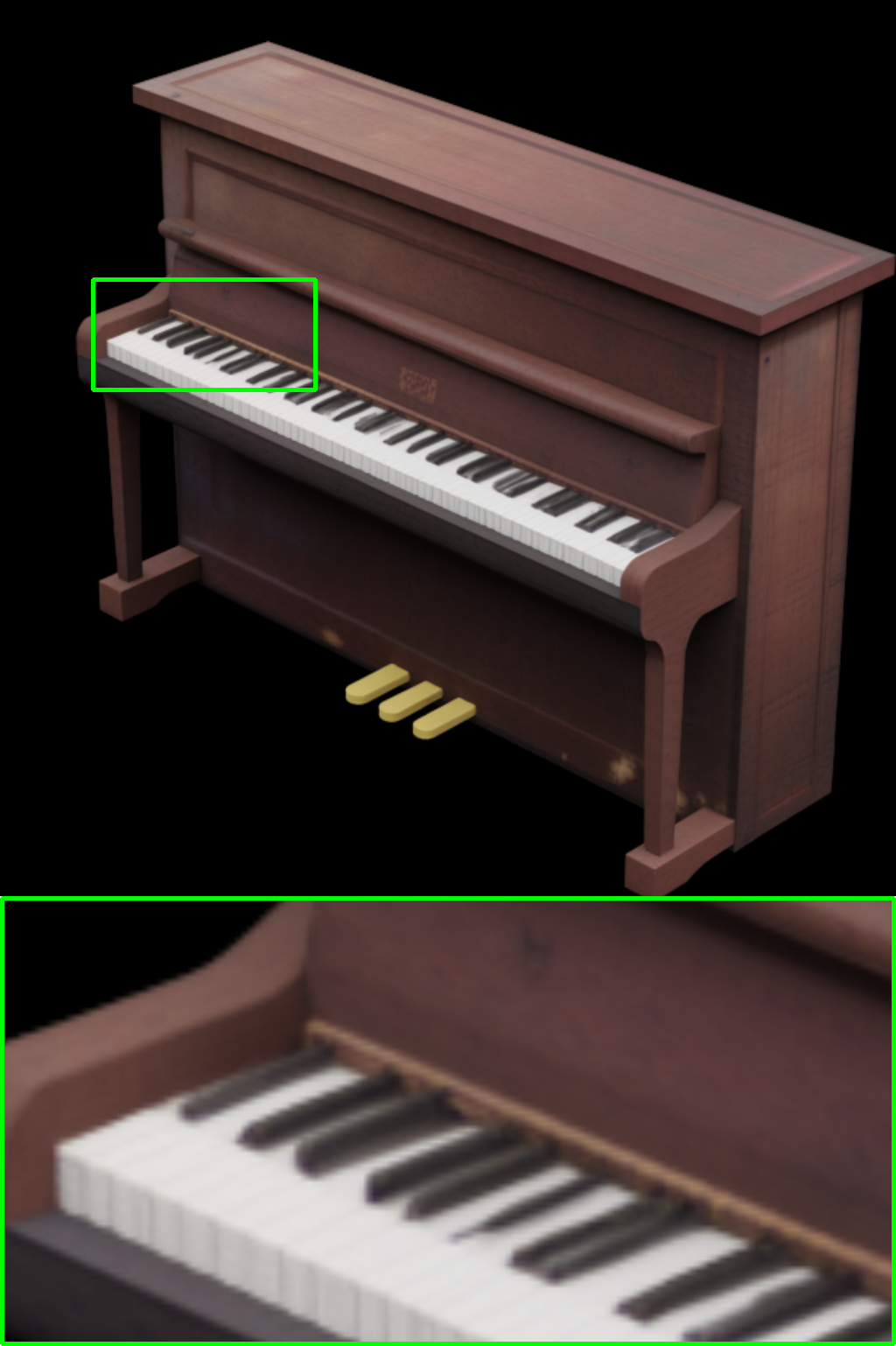}
}
\subfloat{
    \includegraphics[width=0.11\textwidth]{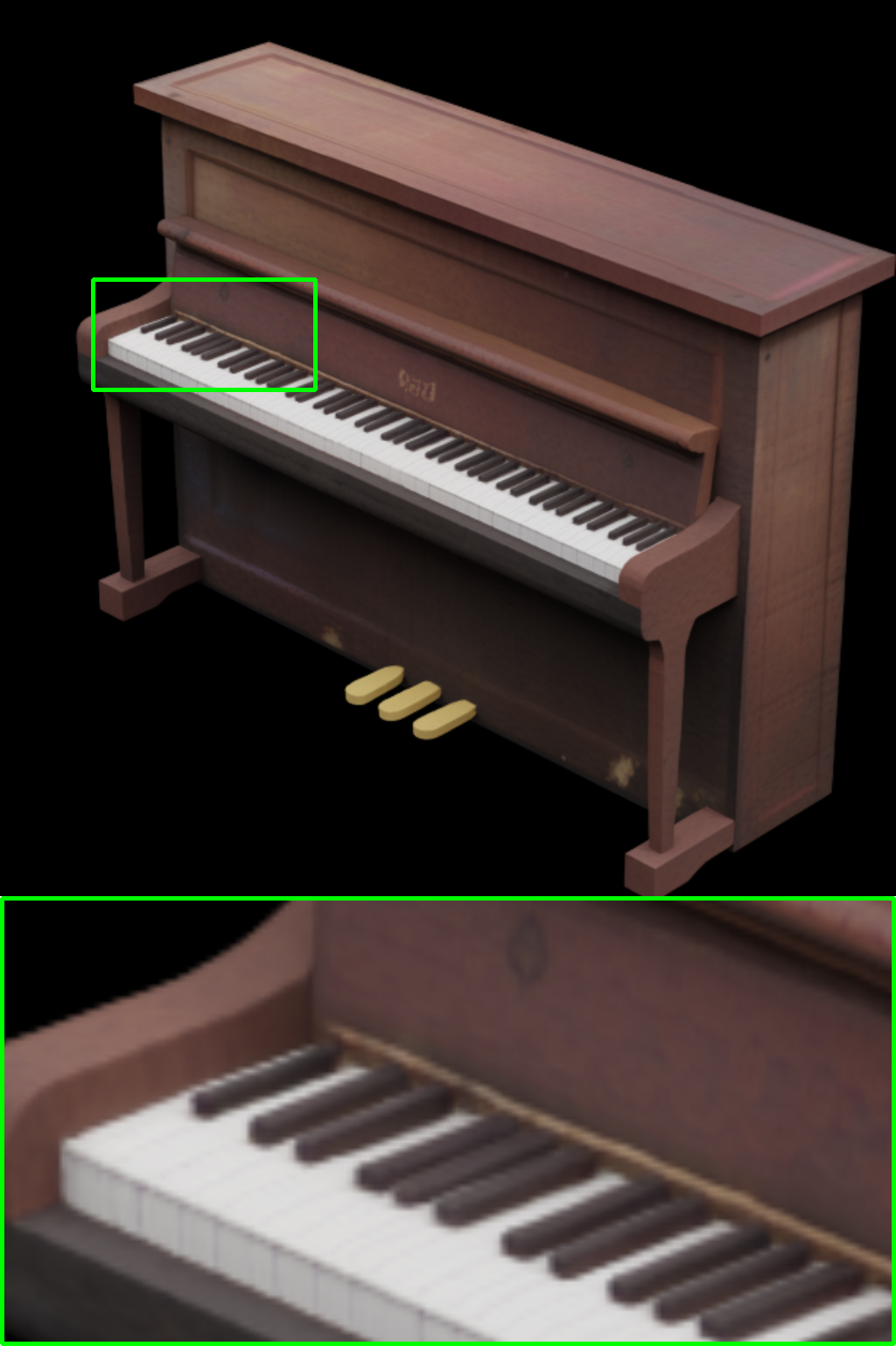}
}

\subfloat{
    \includegraphics[width=0.105\textwidth]{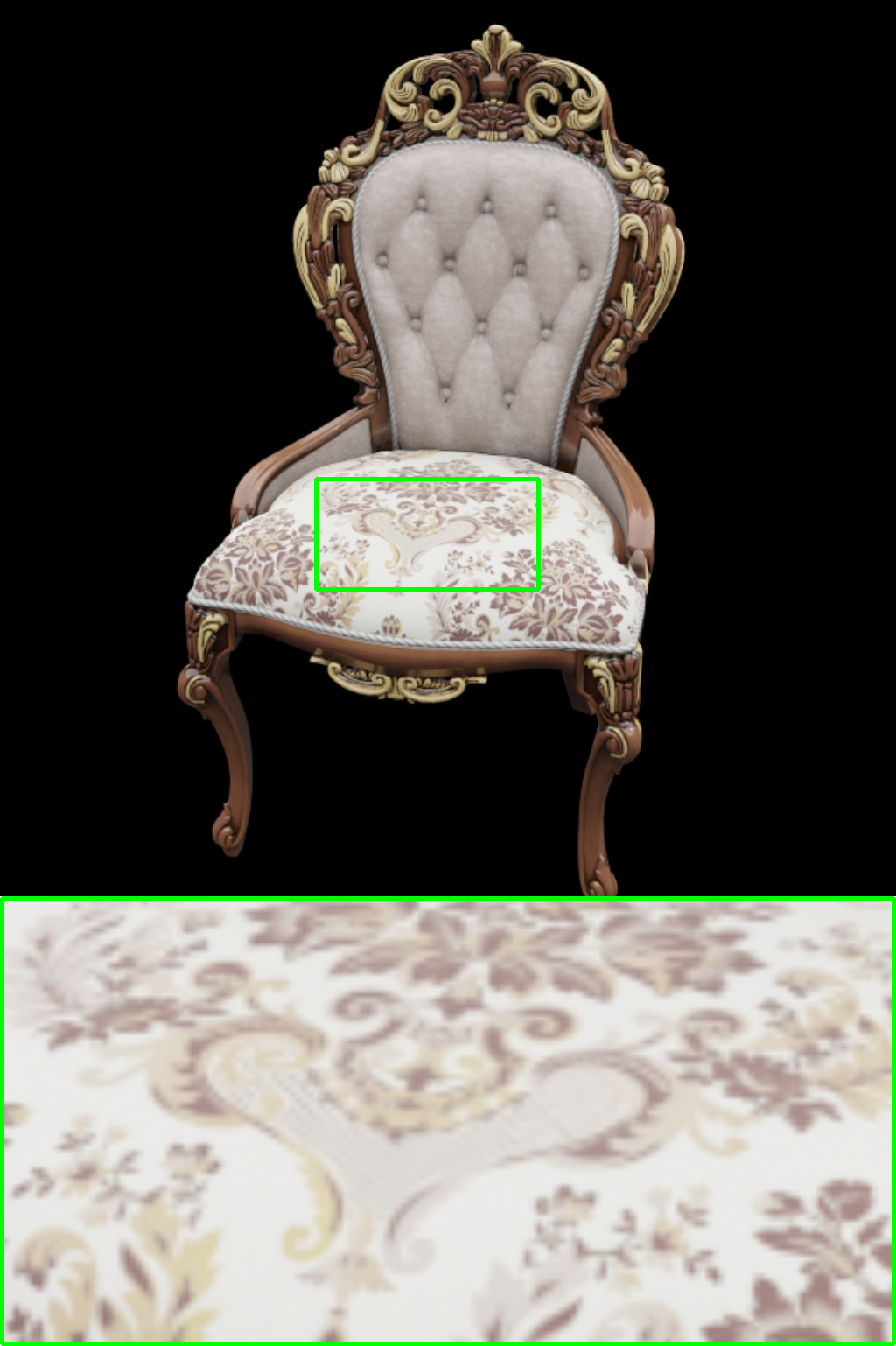} 
}
\subfloat{
    \includegraphics[width=0.11\textwidth]{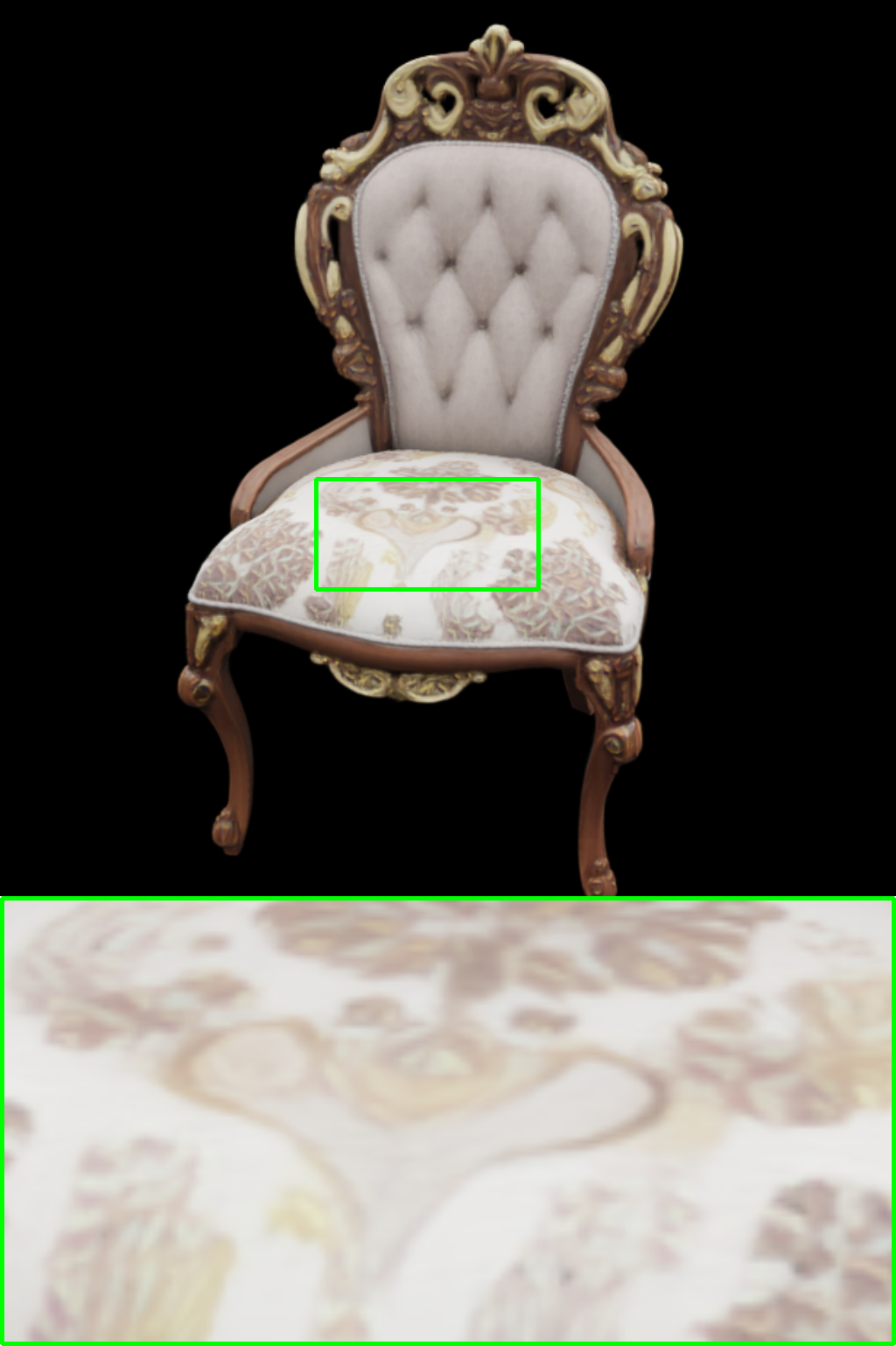}
}
\subfloat{
    \includegraphics[width=0.11\textwidth]{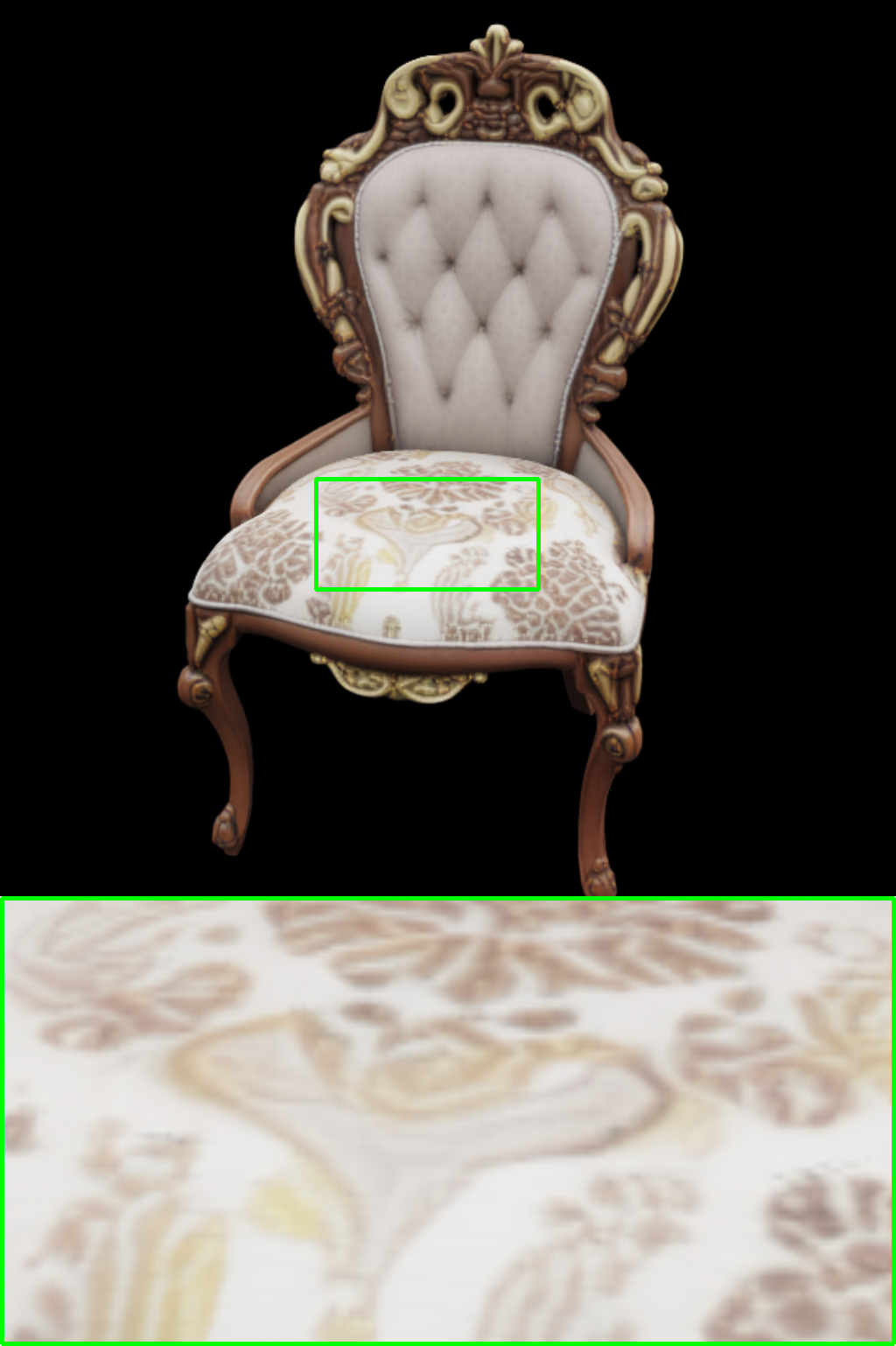}
}
\subfloat{
    \includegraphics[width=0.11\textwidth]{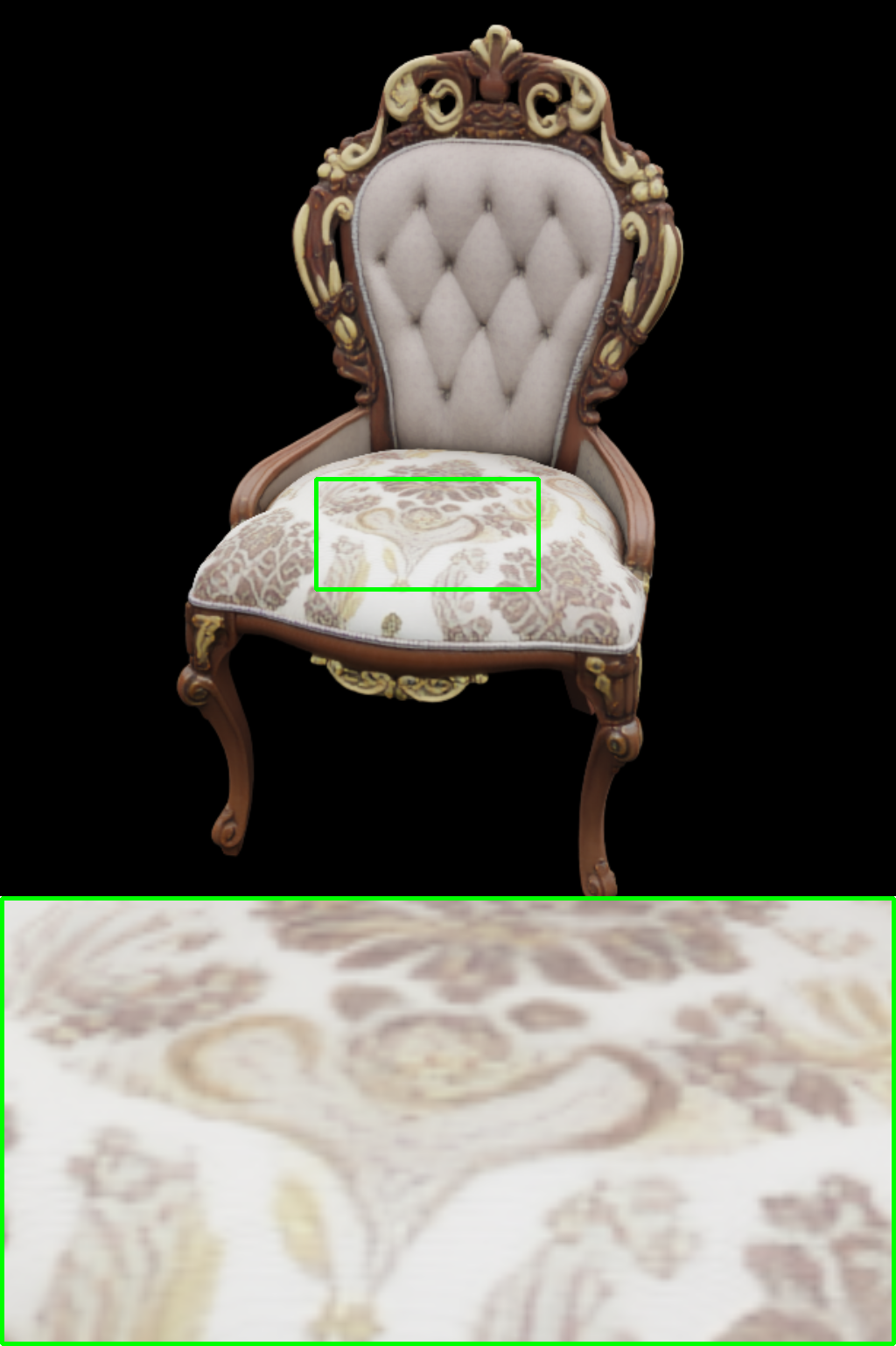}
}

\subfloat{
    \includegraphics[width=0.11\textwidth]{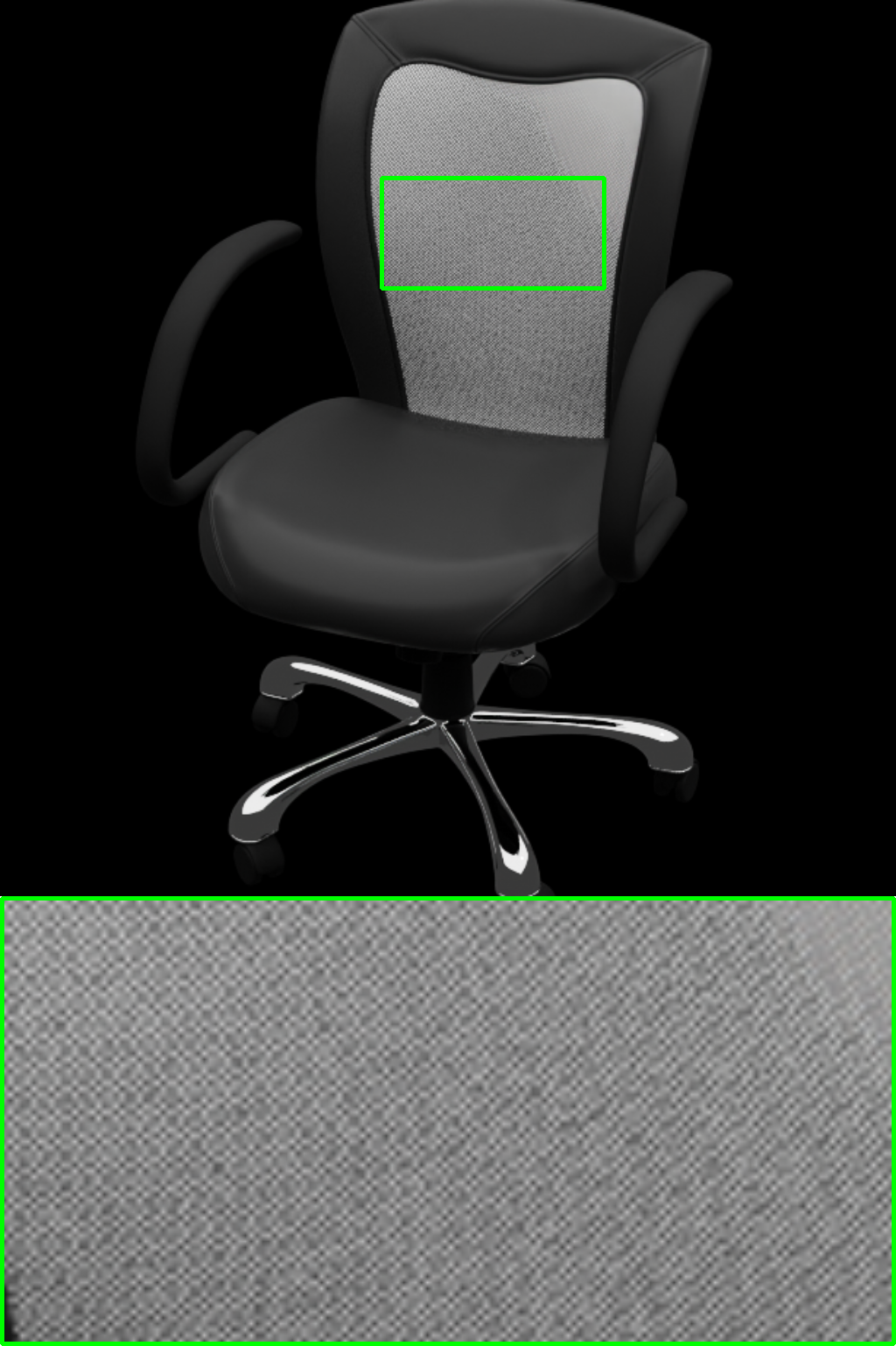} 
}
\subfloat{
    \includegraphics[width=0.11\textwidth]{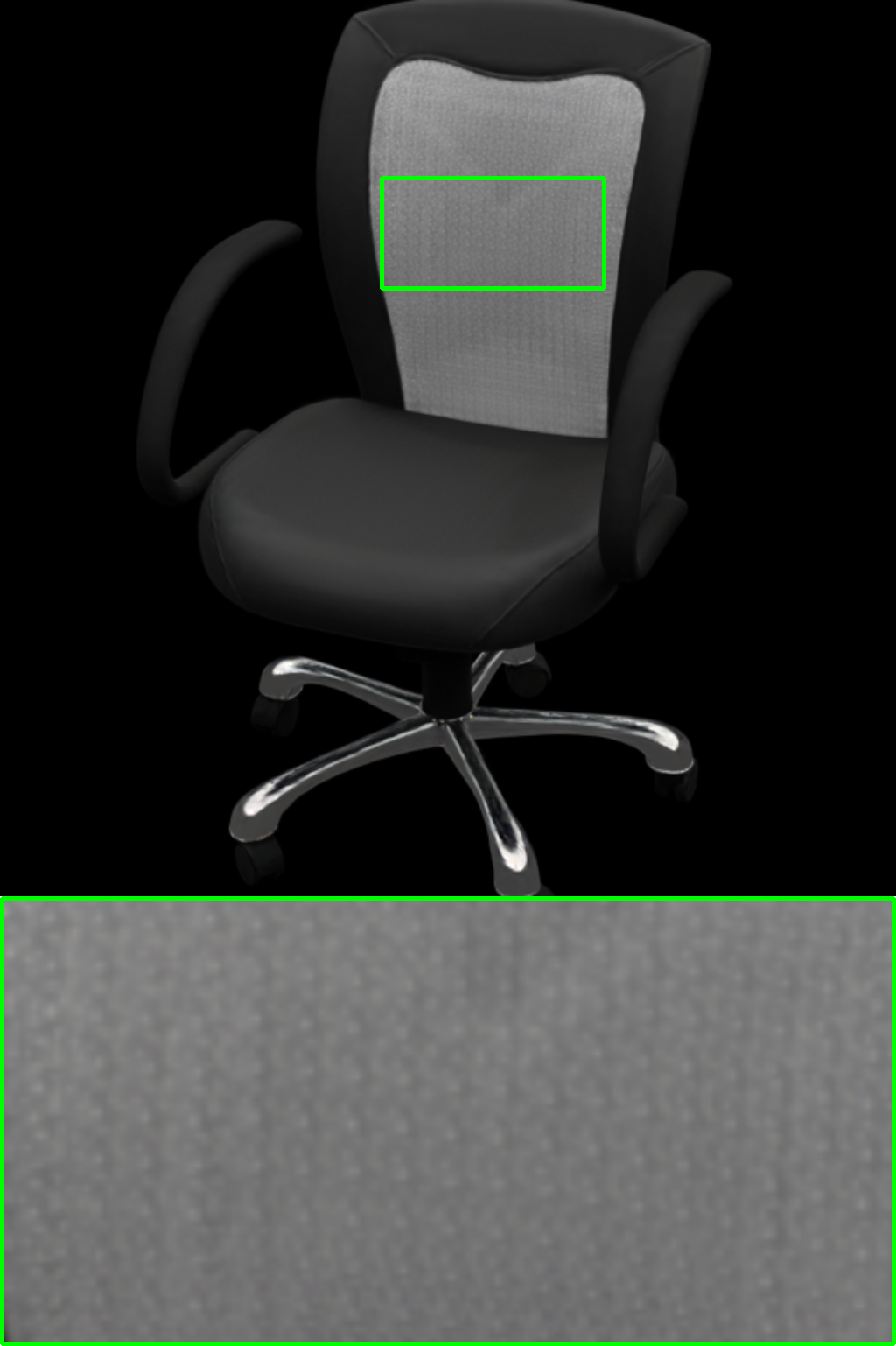}
}
\subfloat{
    \includegraphics[width=0.11\textwidth]{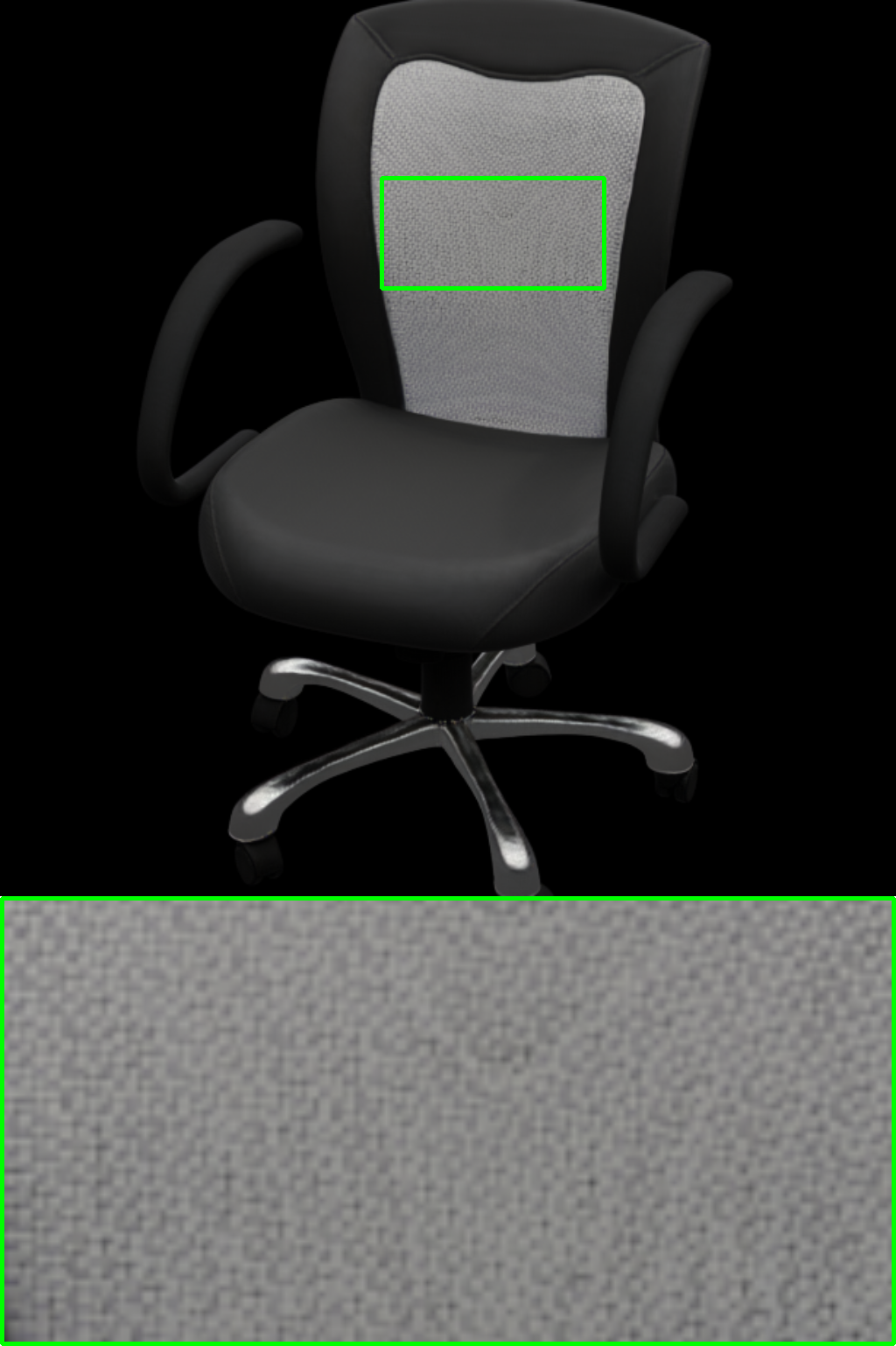}
}
\subfloat{
    \includegraphics[width=0.11\textwidth]{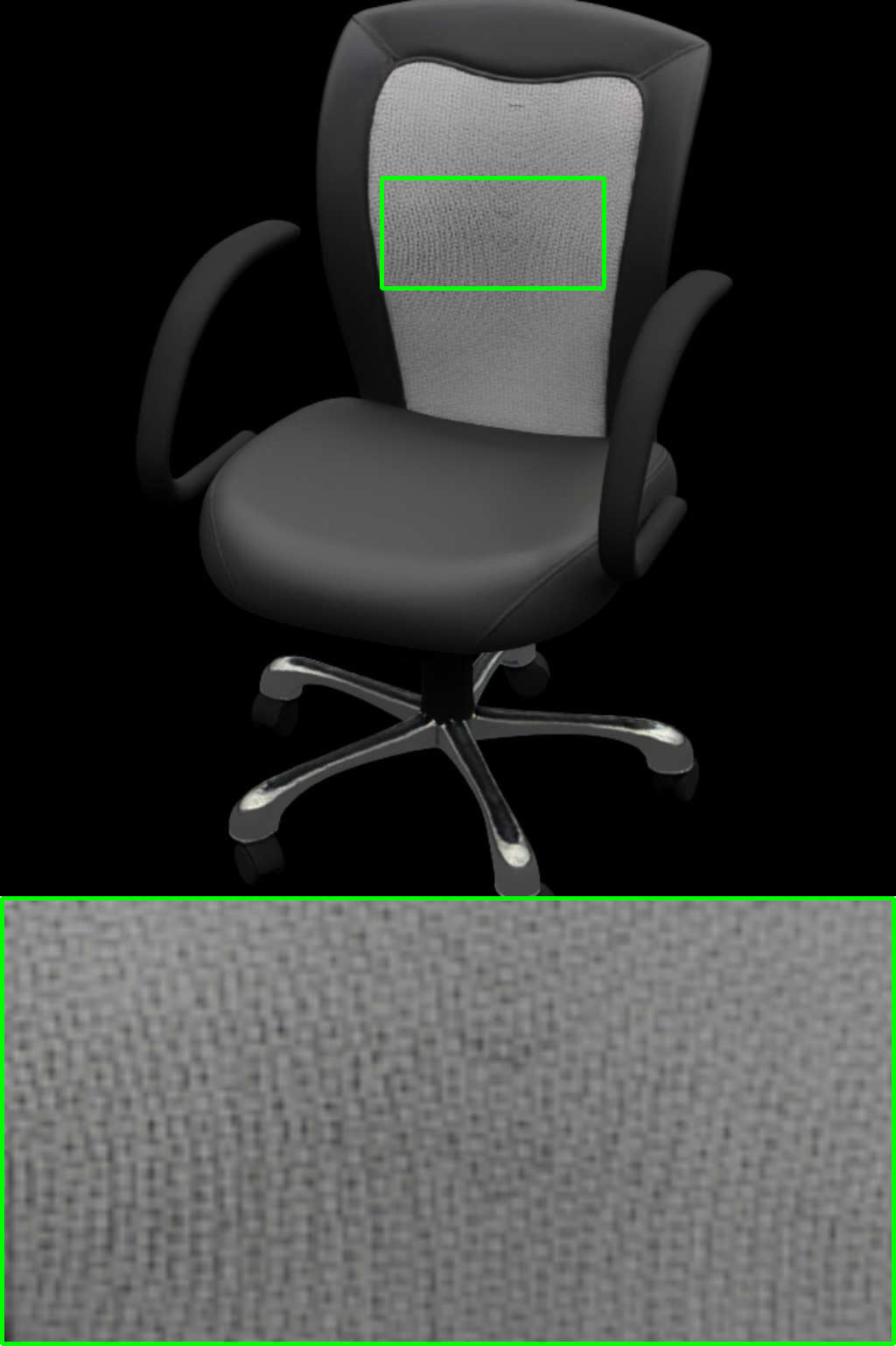}
}

\subfloat[GT]{
    \includegraphics[width=0.11\textwidth]{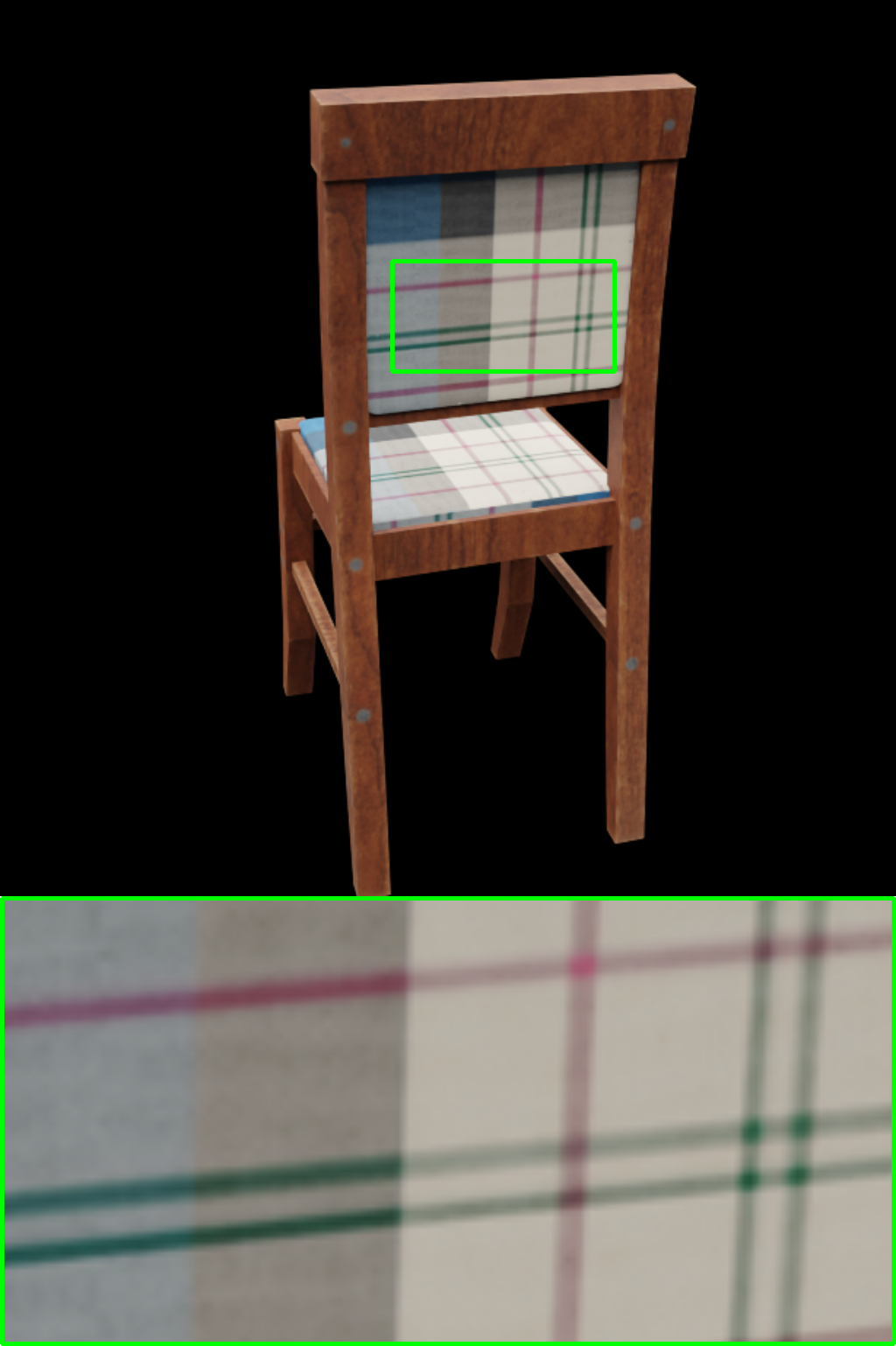} 
}
\subfloat[TRELLIS-64]{
    \includegraphics[width=0.11\textwidth]{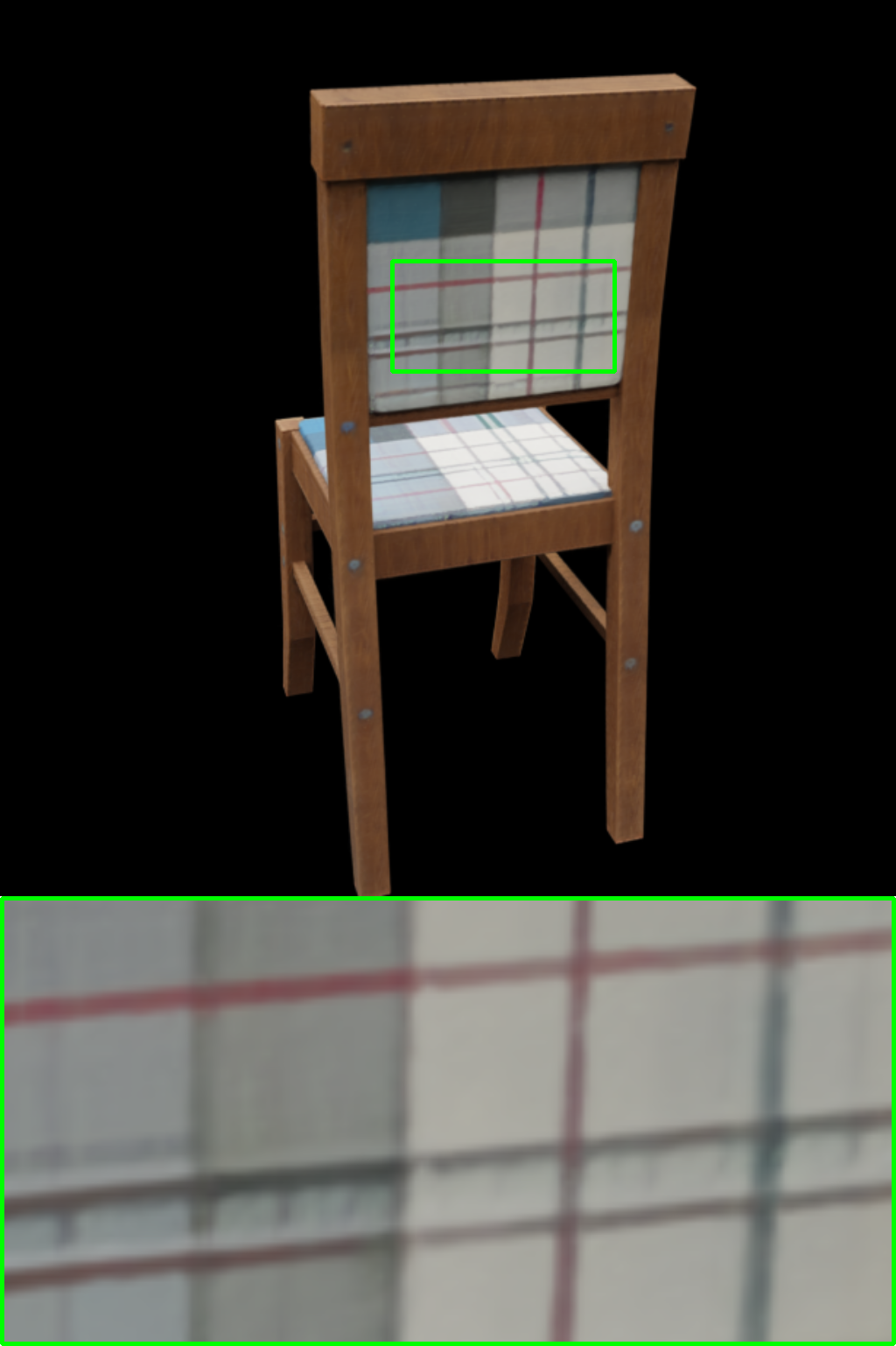}
}
\subfloat[Ours-256]{
    \includegraphics[width=0.11\textwidth]{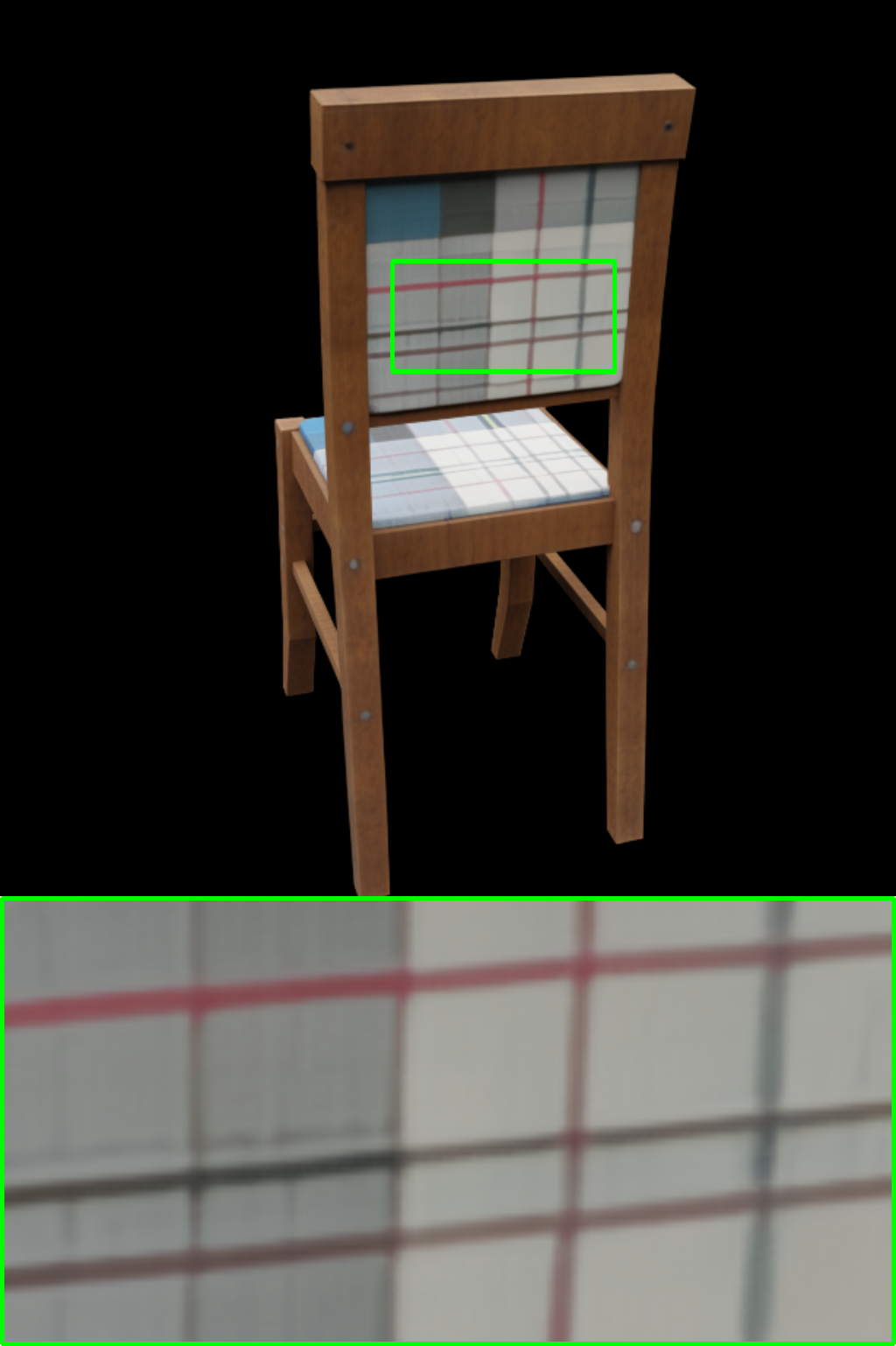}
}
\subfloat[Ours-512]{
    \includegraphics[width=0.11\textwidth]{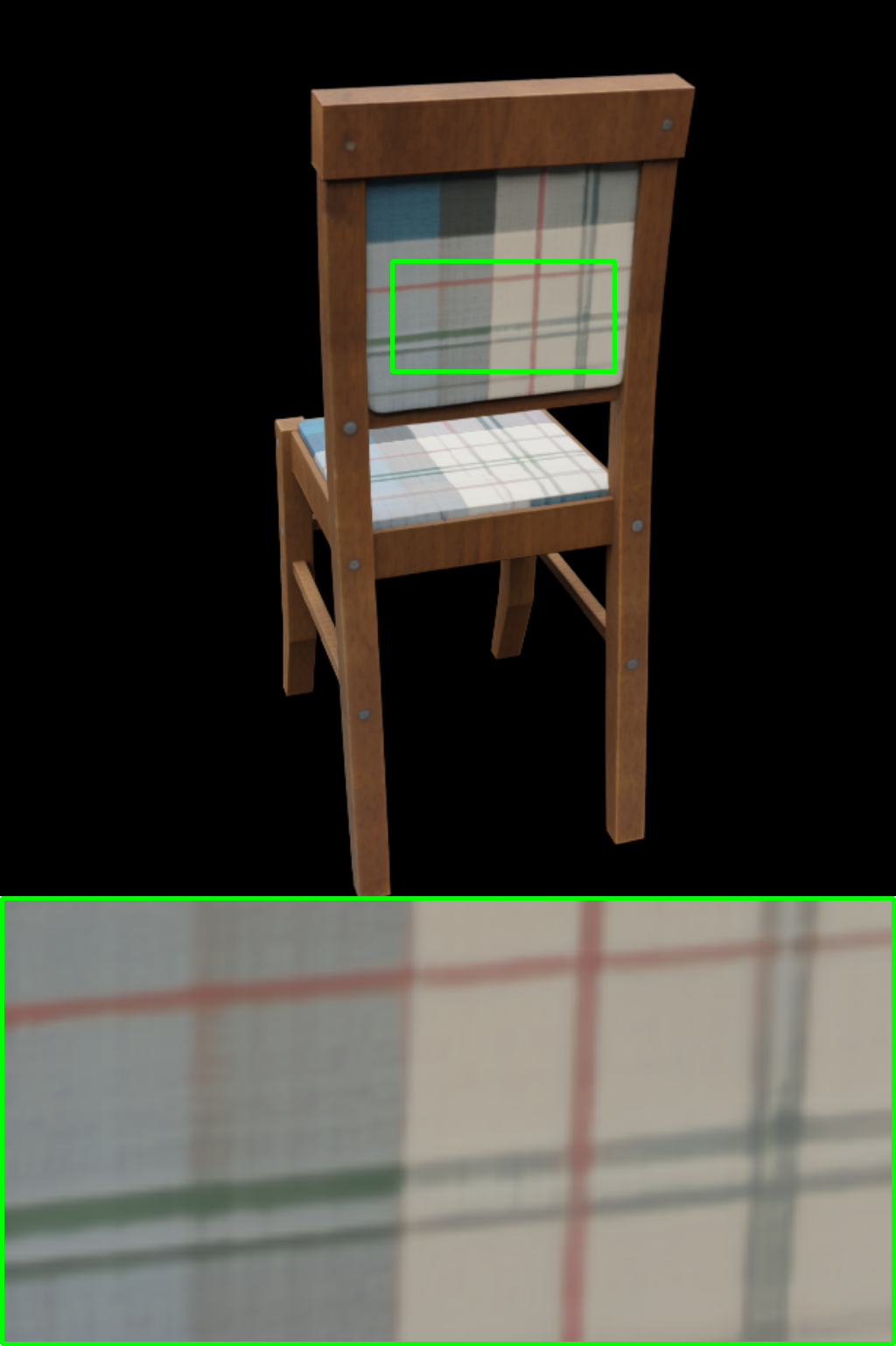}
}
\vspace{-0.8em}
\caption{Qualitative comparison of Texture VAE reconstruction quality between our method with different resolution.}
\label{fig: VAE reconstruction results}
\vspace{-1.2em}
\end{figure}

\subsection{View-domain Partitioning}
\label{sec:View-domain Partitioning}
To curb the rendering memory footprint while preserving high rendering quality, we propose a novel \textit{view-domain partitioning} strategy. It optimizes computational efficiency and memory usage via localized rendering and updates, with the core idea of splitting the rendering process into smaller, manageable segments—this reduces overall memory load and enables more scalable, efficient training.


\textbf{Tile-Based Partitioning Strategy.} The first step is partitioning the valid image region into uniformly sized, overlapping tiles. Tile overlap is crucial—it avoids boundary discontinuities and artifacts, enabling smooth transitions and high-quality rendering across the entire image region.


During training, tiles are randomly sampled from the partitioned image region, introducing stochasticity that reduces overfitting and improves generalization. Each selected tile then undergoes culling to remove voxels outside its camera frustum, which minimizes memory usage by retaining only relevant voxels for rendering.


\textbf{Localized Rendering and Updates.} The remaining voxels in the selected tile generate 3D Gaussians for localized rendering of the corresponding image patch. Losses are computed and gradients back-propagated only within this patch. Focusing on one patch at a time cuts computational load and memory needs vs. rendering the whole image at once, enabling high-quality rendering with practical memory usage.

\subsection{Loss Function}
We use the binary cross-entropy loss to supervise the mask generated by the Mask Generator for redundancy removal:
\begin{equation}
    L_{mask} = BCE(M, \hat{M}),
\end{equation}

We train the Texture-VAE in an end-to-end manner, with the objective function as follows\cite{kerbl20233d,xiang2025structured}:
\begin{equation}
    L = \lambda_1 \mathcal{L}_{1} + \lambda_2 \mathcal{L}_{SSIM} +\lambda_3 \mathcal{L}_{LPIPS} + \lambda_4 \mathcal{L}_{\text{KL}} + \lambda_5 \mathcal{L}_{reg},
\end{equation}
We calculate the L1, D-SSIM, and LPIPS losses between the ground truth (GT) and the 3DGS rendered images output by the Texture-VAE. $\mathcal{L}_{KL}$ is the KL divergence between the learned latent distribution and a standard normal prior, regularizing the latent space. $\mathcal{L}_{reg}$ is a regularization term for the opacity and scale properties of the output 3DGS, which encourages small and opaque Gaussians.

\section{EXPERIMENTS}
\label{sec:exp}
\subsection{Experiment Settings}


\textbf{Datasets.} We selected a total of 120k high-quality data samples from three datasets, namely Objaverse (XL)\cite{deitke2023objaverse}, ABO\cite{collins2022abo}, and 3D-FUTURE\cite{fu20213d}, for training the Texture Variational Autoencoder (VAE) and Latent Space Flow models. Specifically, for the Objaverse (XL)\cite{deitke2023objaverse} dataset, we used the intersection of the filtered results obtained via the TRELLIS\cite{xiang2025structured} and Step1X-3D\cite{li2025step1x} methods. We employed Toys4K\cite{stojanov2021using} as our test set.

\textbf{Baselines.} We compare our Texture-VAE at different voxel resolutions with TRELLIS\cite{xiang2025structured}. For the generated results, we also compare with TRELLIS\cite{xiang2025structured}, hunyuan3D 2.1\cite{hunyuan3d2025hunyuan3d} and Step1X-3D\cite{li2025step1x}.

\textbf{Metrics.} We use PSNR, LPIPS, and SSIM to evaluate texture reconstruction loss. We use FID\cite{parmar2022aliased} and KID\cite{binkowski2018demystifying} to evaluate texture generation quality.

\subsection{VAE Reconstruction Evaluation}
We conducted a comparative analysis on the reconstruction quality of Texture Variational Autoencoders (VAEs) under different voxel resolutions, with experiments carried out on the original TRELLIS-64, Ours-256, and Ours-512. As shown in tab.\ref{tab: Quantitative comparison of reconstruction results}, with the gradual improvement of voxel resolution, the texture reconstruction quality exhibits a significant positive growth trend. Under the 512-resolution condition, the texture reconstruction effect reaches the optimal level.

Combined with the qualitative results in fig.\ref{fig: VAE reconstruction results}, we further found that high-resolution voxelization can capture the detailed features of textures more accurately: the texture details are richer and more realistic, and the gap between the reconstructed images and the original ones is significantly reduced. This result strongly demonstrates the key role of high-resolution voxels in improving the quality of texture reconstruction.
\begin{table}[t]
    \vspace{-0.5em}
    \renewcommand{\arraystretch}{1.2}
    \setlength{\tabcolsep}{12pt}
    \centering
    \caption{\upshape{\textbf{Quantitative comparison of Texture VAE reconstruction quality}  between our method with different resolution and TRELLIS.}}
    \vspace{-0.5em}
    \label{tab: Quantitative comparison of reconstruction results}
    \begin{tabular}{c|ccc}
        \hline
         & LPIPS $\downarrow$ & PSNR $\uparrow$ & SSIM $\uparrow$  \\
        \hline
        TRELLIS-64 &  0.0219 & 34.26 & 0.984  \\
        Ours-256 & 0.0195 & 34.49 & 0.984 \\
        Ours-512 &\textbf{0.0187} & \textbf{34.97} & \textbf{0.985} \\
        \hline
    \end{tabular}
    \vspace{-0.5em}
\end{table}

\begin{table}[t]
    \renewcommand{\arraystretch}{1.2}
    \setlength{\tabcolsep}{14pt}
    \centering
    \caption{\upshape{\textbf{Quantitative generation results on Toys4k.}}}
    \vspace{-0.5em}
    \begin{tabular}{c|cc}
        \hline
         & FID $\downarrow$ & KID $\downarrow$   \\
        \hline
        TRELLIS &  83.496 & $5.760\times10^{-4}$ \\
        hunyuan3D 2.1 & 74.219 & $4.988\times10^{-4}$ \\
        Step1X-3D & 99.620 & $28.36\times10^{-4}$\\
        Ours & \textbf{67.766} & $\mathbf{3.647\times10^{-4}}$\\
        \hline
    \end{tabular}
    \vspace{-1.5em}
    \label{tab: 3D generation results}
\end{table}

\subsection{Image to 3D Generation}
\begin{figure}[t]
\captionsetup[subfloat]{labelsep=none,format=plain,labelformat=empty,font=scriptsize}
\centering
\subfloat{
    \includegraphics[width=0.085\textwidth]{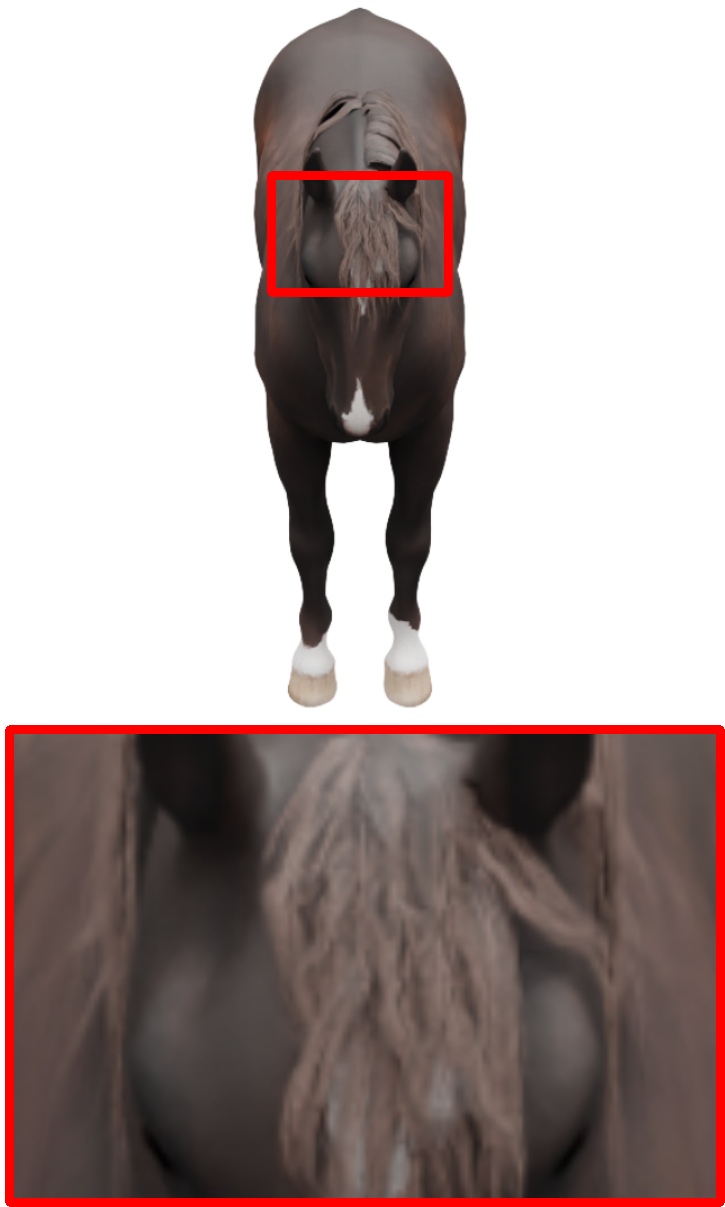} 
}
\subfloat{
    \includegraphics[width=0.085\textwidth]{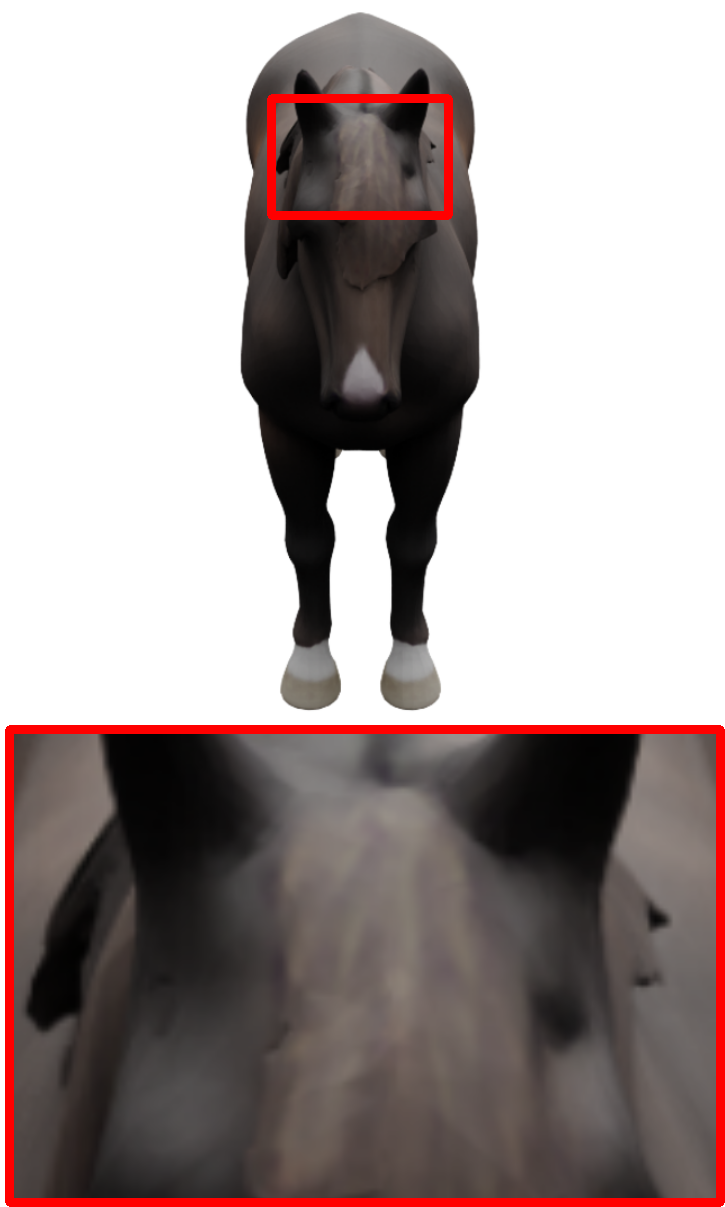}
}
\subfloat{
    \includegraphics[width=0.085\textwidth]{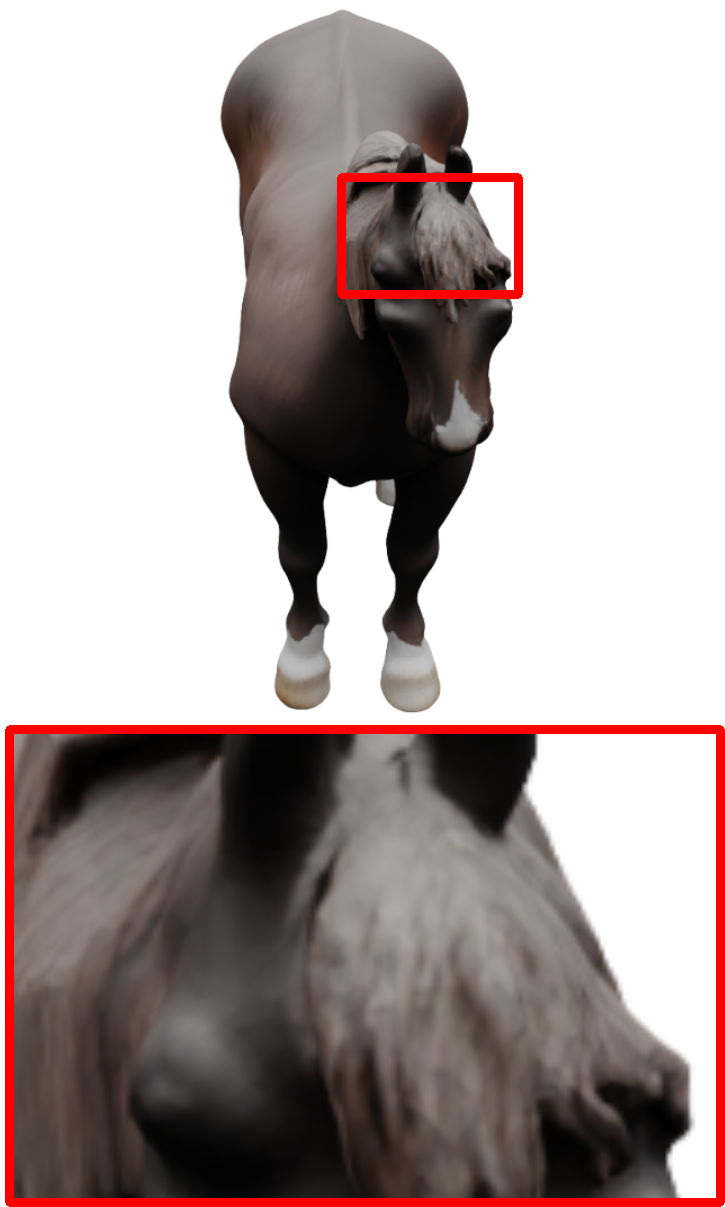}
}
\subfloat{
    \includegraphics[width=0.085\textwidth]{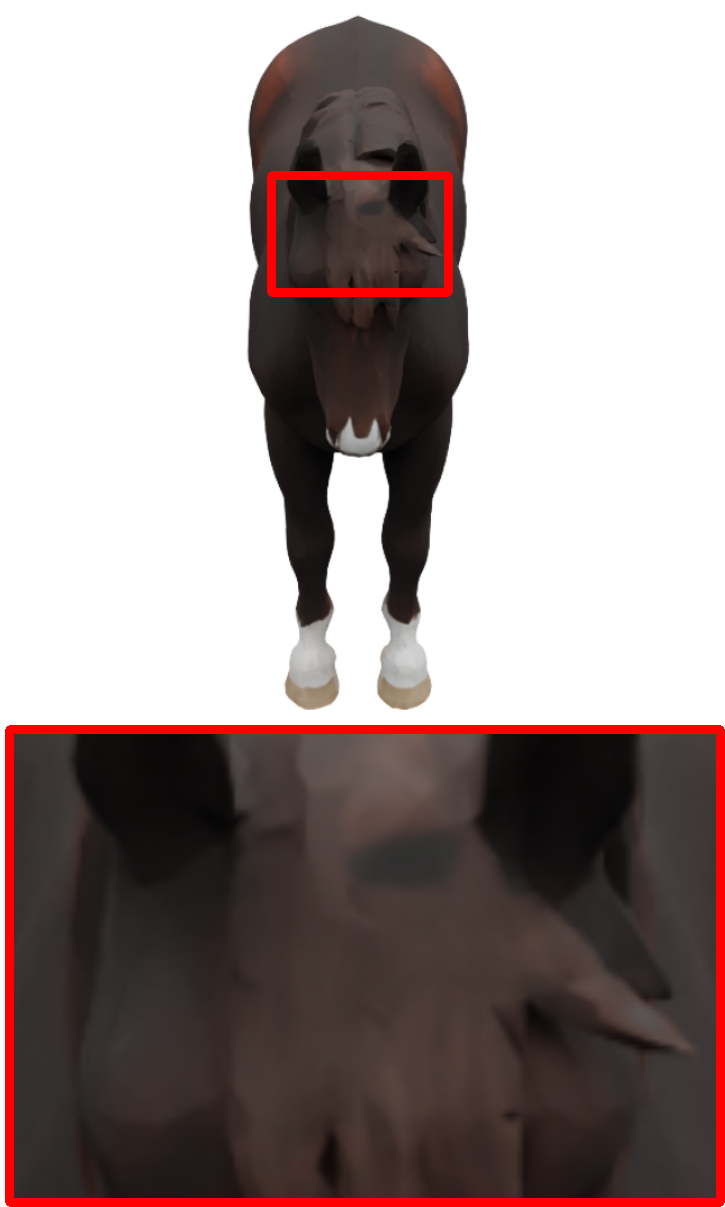}
}
\subfloat{
    \includegraphics[width=0.085\textwidth]{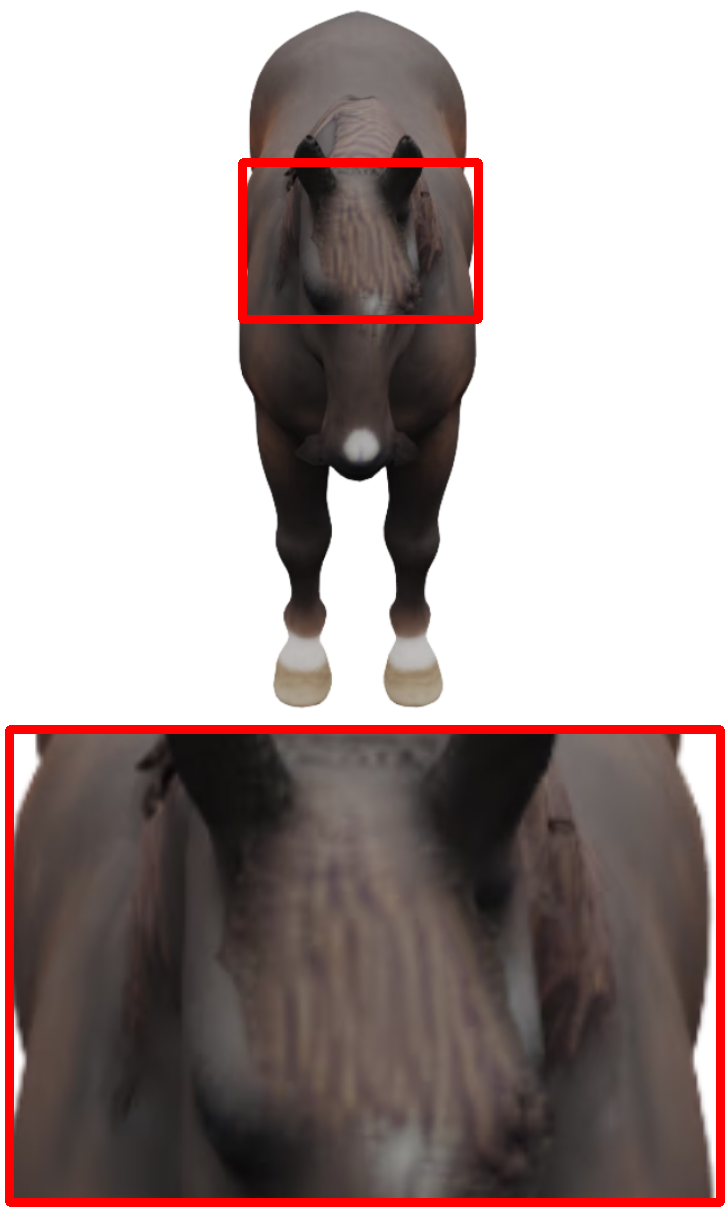}
}

\subfloat{
    \includegraphics[width=0.085\textwidth]{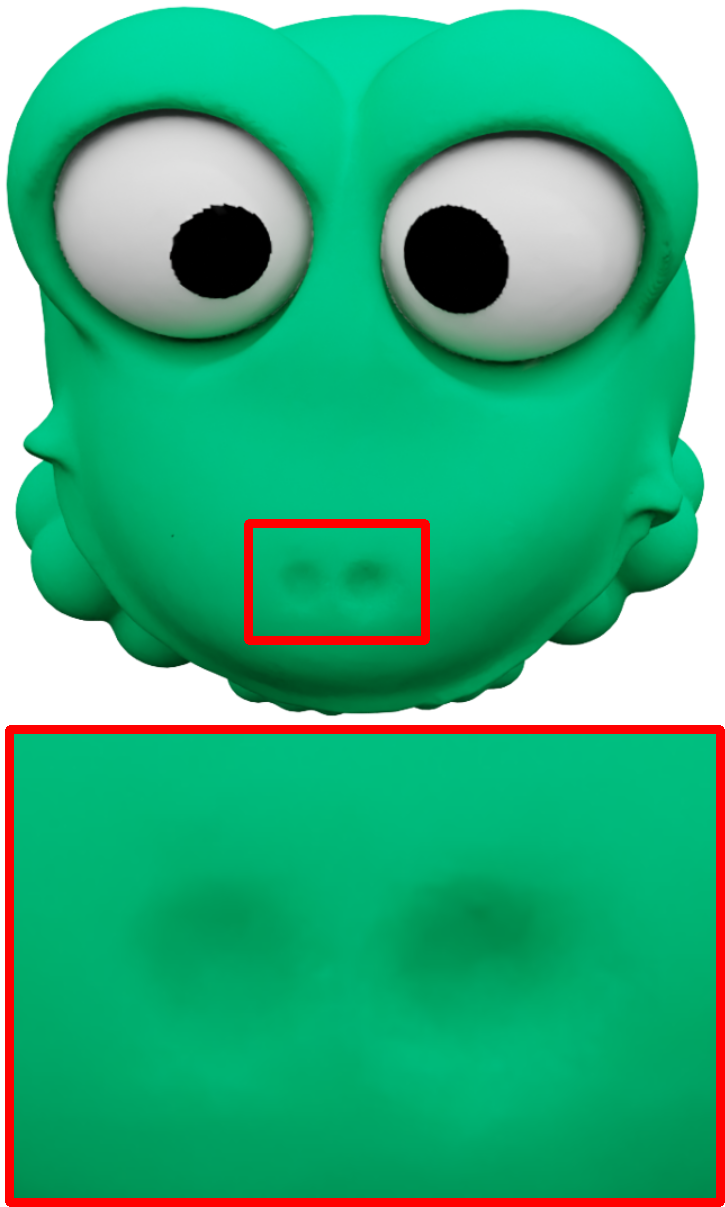} 
}
\subfloat{
    \includegraphics[width=0.085\textwidth]{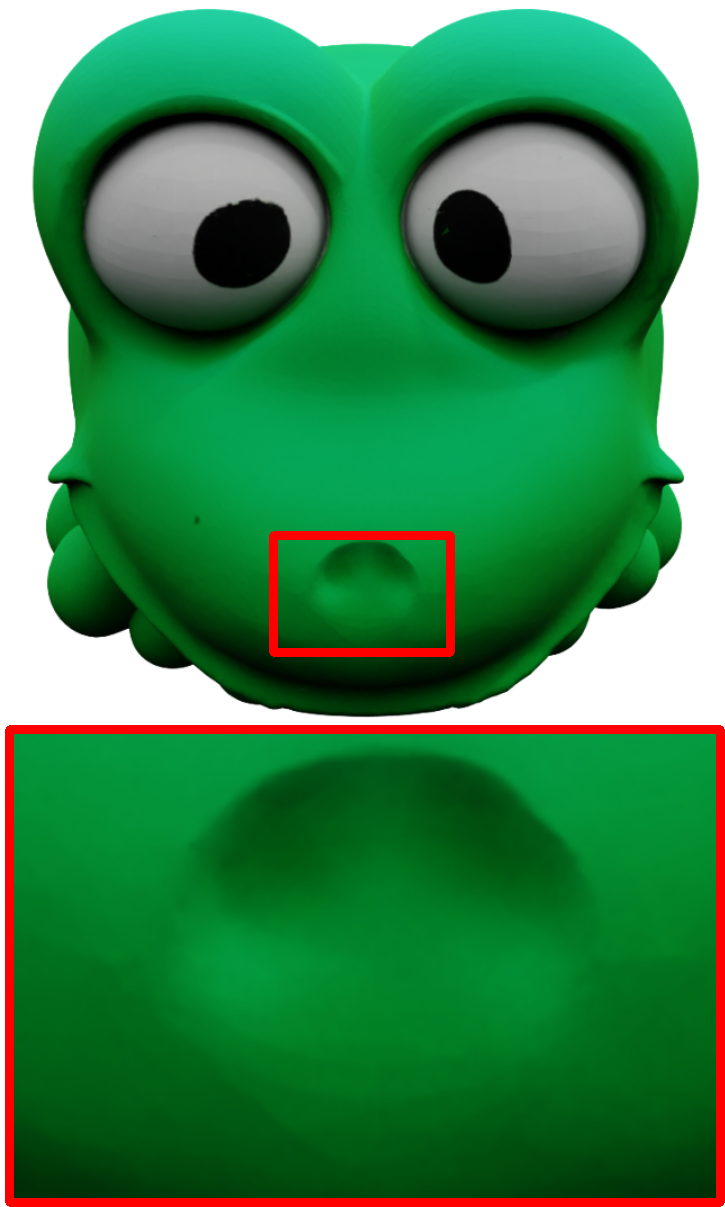}
}
\subfloat{
    \includegraphics[width=0.085\textwidth]{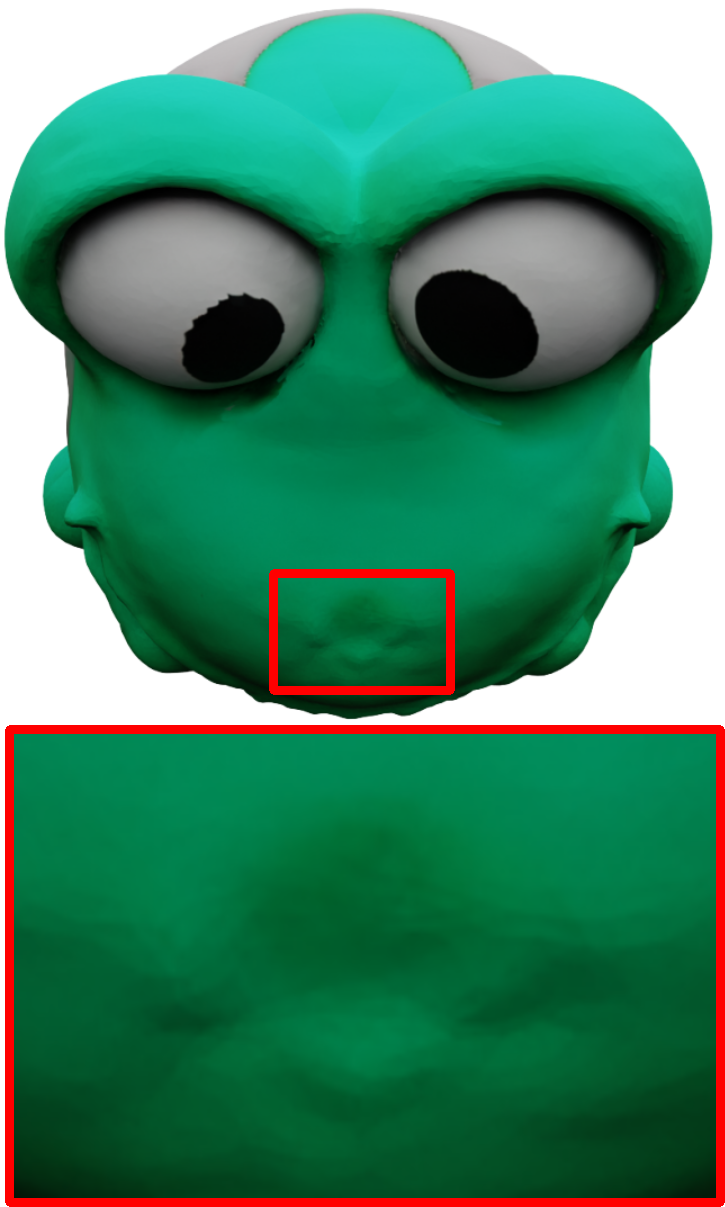}
}
\subfloat{
    \includegraphics[width=0.085\textwidth]{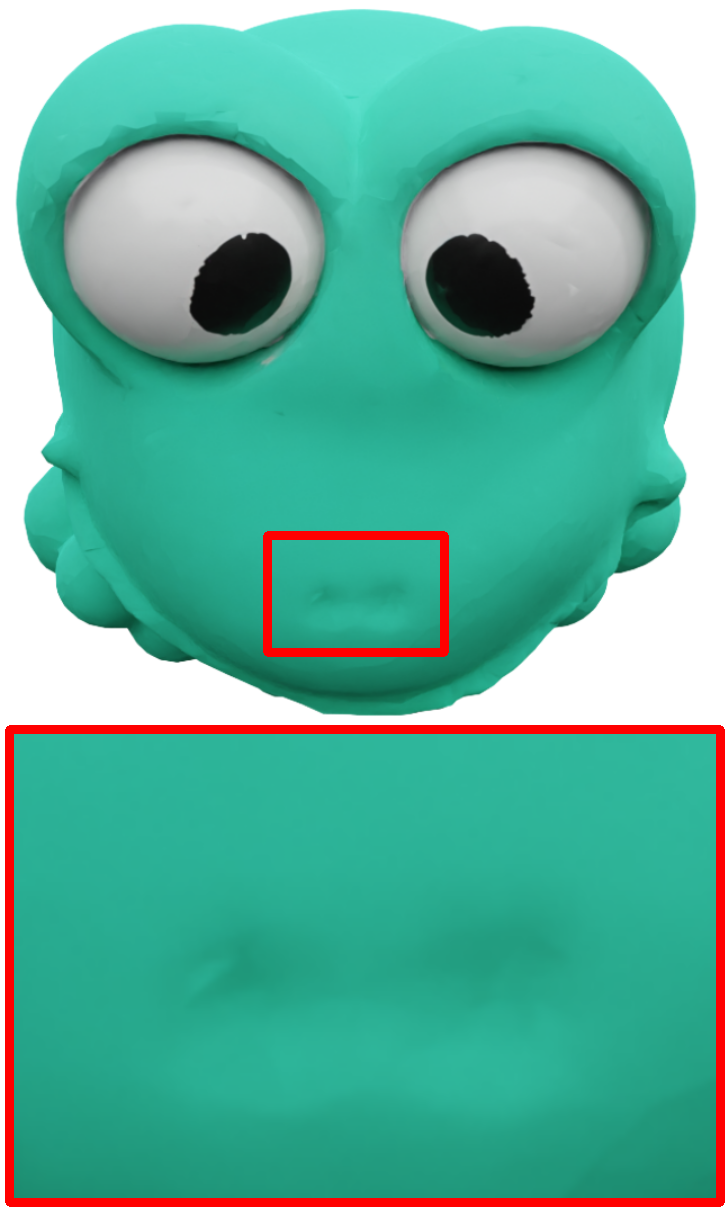}
}
\subfloat{
    \includegraphics[width=0.085\textwidth]{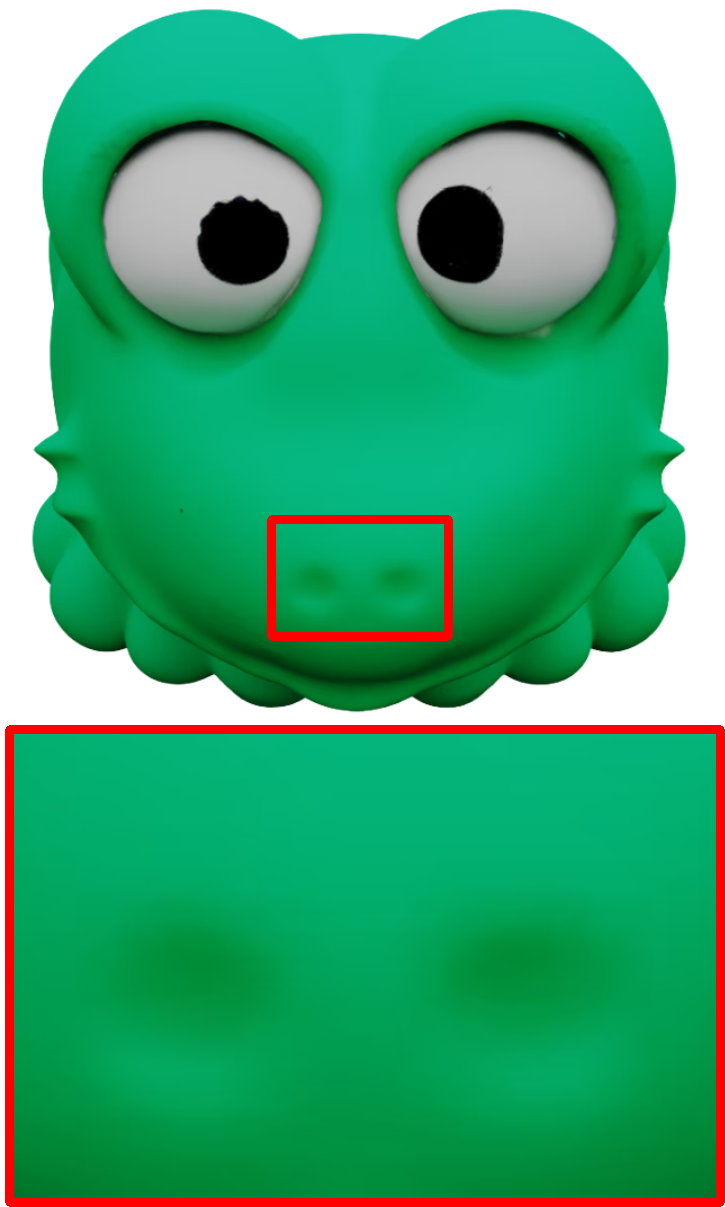}
}

\subfloat[GT]{
    \includegraphics[width=0.085\textwidth]{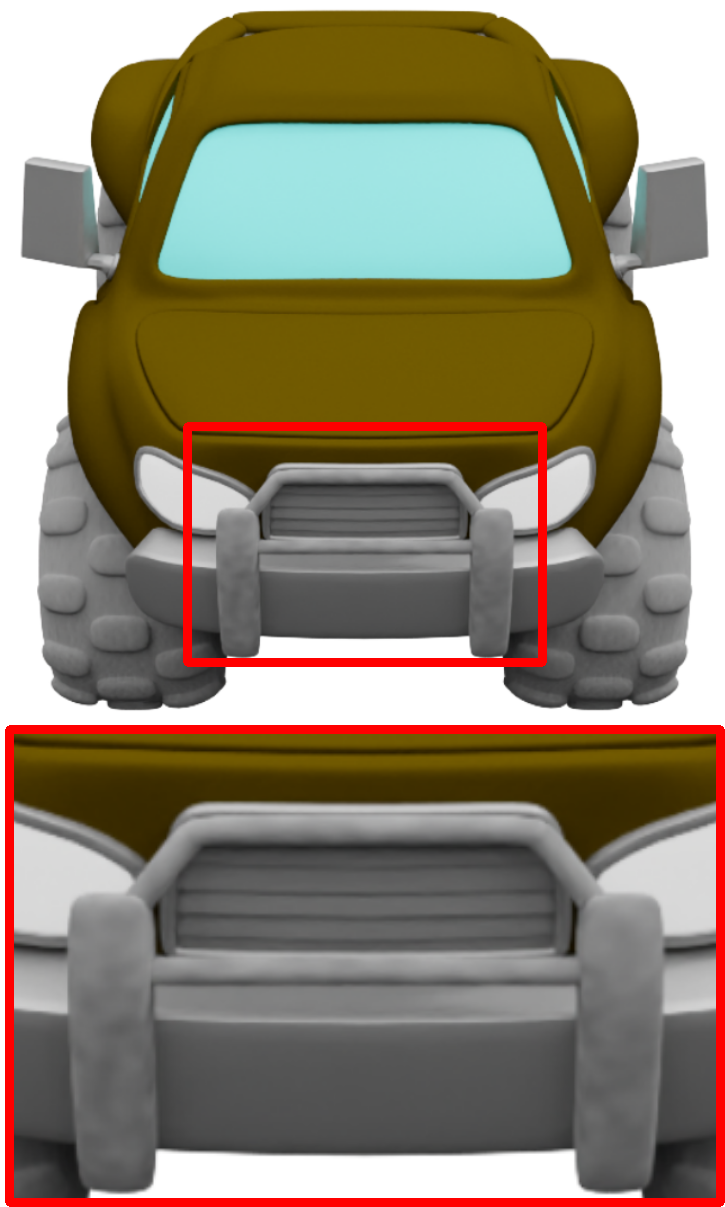} 
}
\subfloat[TRELLIS]{
    \includegraphics[width=0.085\textwidth]{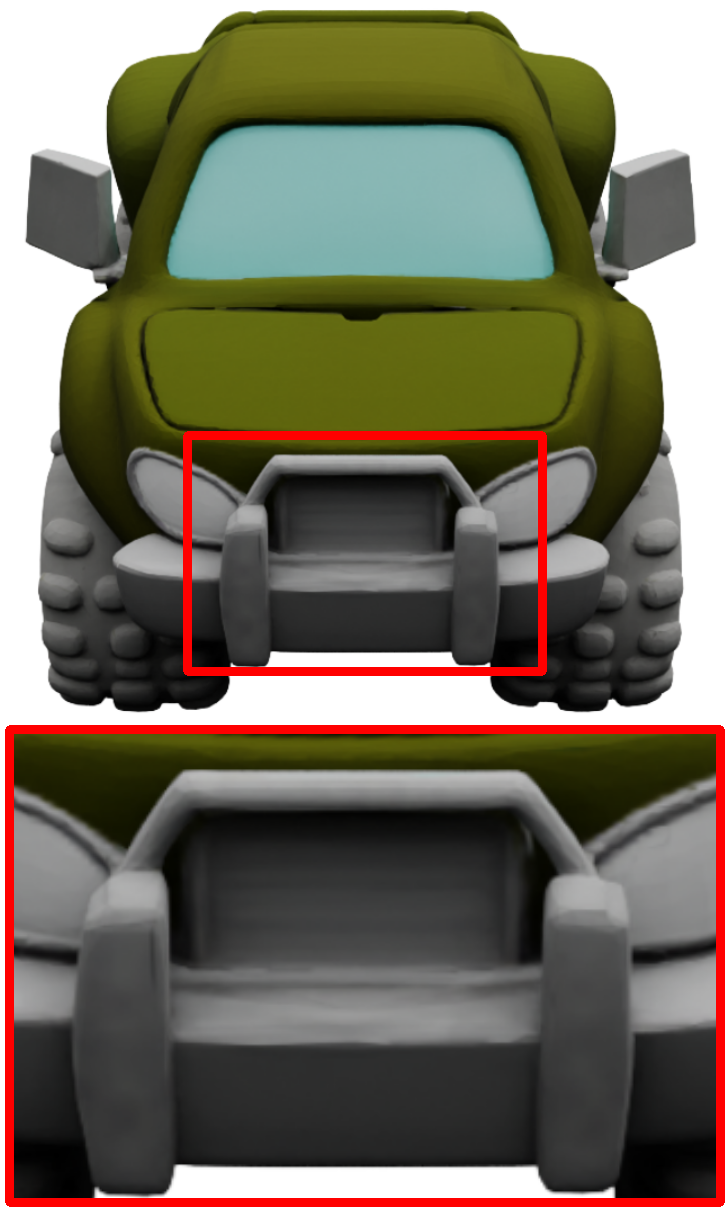}
}
\subfloat[Step1X-3D]{
    \includegraphics[width=0.085\textwidth]{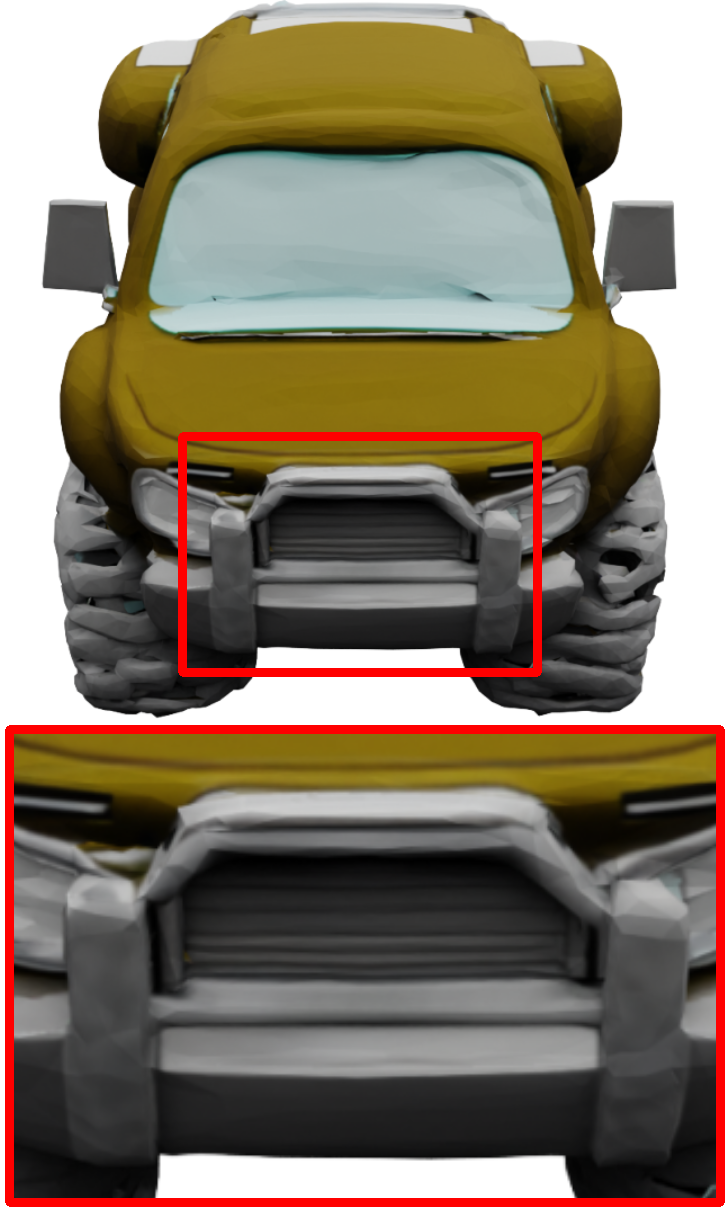}
}
\subfloat[hunyuan3D 2.1]{
    \includegraphics[width=0.085\textwidth]{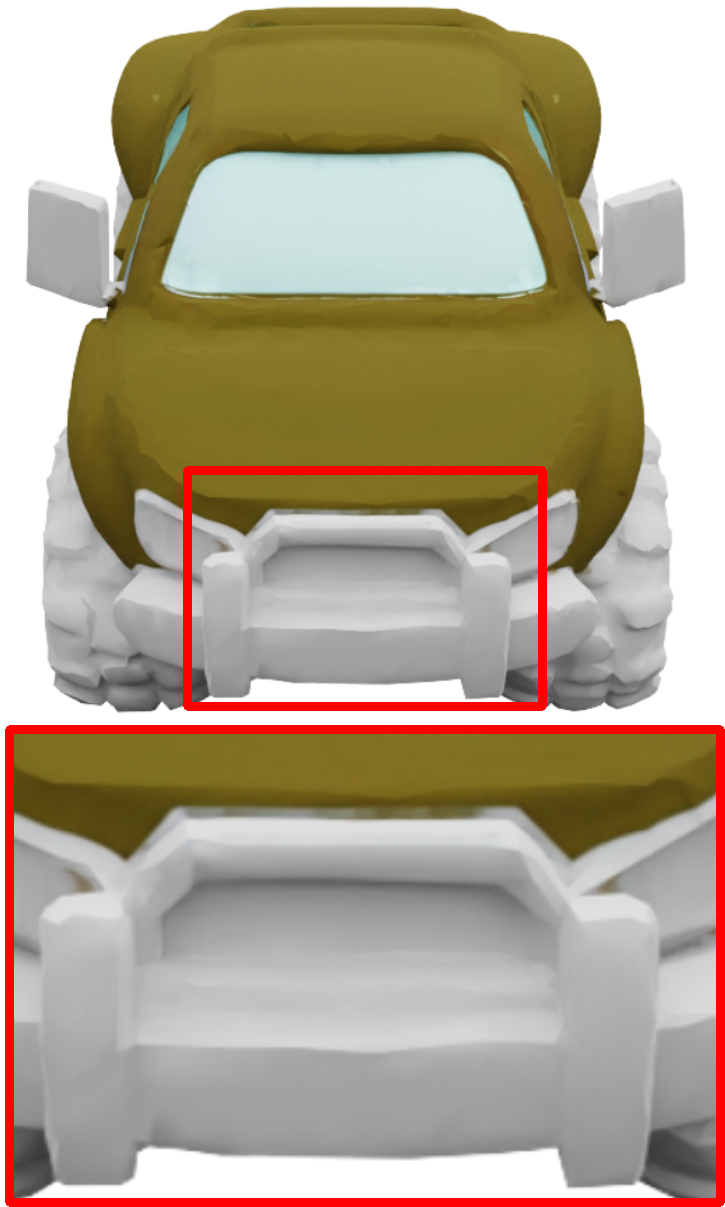}
}
\subfloat[ours]{
    \includegraphics[width=0.085\textwidth]{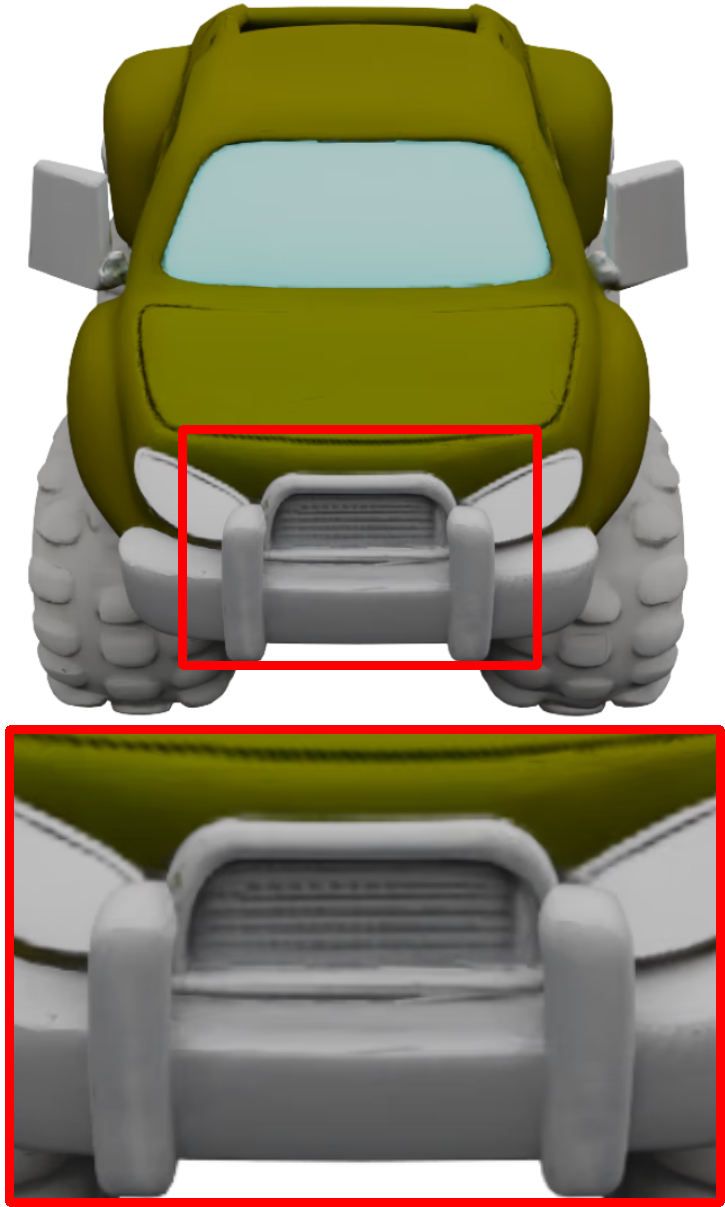}
}
\vspace{-0.5em}
\caption{Qualitative comparison of image to 3D generation.}
\label{fig: 3D generation results.}
\vspace{-1.2em}
\end{figure}

We have also verified the effectiveness of the Texture-VAE as a foundation model for generation tasks. Tab.\ref{tab: 3D generation results} confirms the validity of our generated results, while the fig.\ref{fig: 3D generation results.} presents the visualization results of 3D generation by different models. The results show that the generated textures preserve fine details and are highly consistent with the ground-truth images, demonstrating the generality of our method.

\subsection{Ablation Studies}
We conducted a detailed evaluation of the GPU memory consumption of the generated mask (with redundant points removed during upsampling) and the view-domain partitioning training under different resolutions on an 80GB H100. The relevant results are presented in Table \ref{tab: evaluation on the GPU memory consumption}. Experimental results show that both the redundant-free upsampling strategy and the view-domain partitioning strategy can significantly reduce GPU memory consumption during the training process, highlighting the efficiency and advantages of our proposed strategies in handling high-resolution data.

\begin{table}[h]
    \renewcommand{\arraystretch}{1.2}
    \setlength{\tabcolsep}{12pt}
    \centering
    \caption{\upshape{\textbf{Comparison of GPU VRAM usage.} Unit: MB.}}
    \vspace{-0.5em}
    \label{tab: evaluation on the GPU memory consumption}
    \begin{tabular}{c|cc}
        \hline
         & res-256 & res-512 \\
        \hline
        raw & 73011 & OOM   \\
        mask & 41577 & 65537   \\
        mask + block $2\times2$ & 27881 & 45811   \\
        mask + block $4\times4$ & 24541 & 38301  \\
        \hline
    \end{tabular}
    \vspace{-1.5em}
\end{table}

\section{CONCLUSIONS}
We have surmounted the constraints of voxel resolution and accomplished high-fidelity texture modeling. Going forward, we will continue to pursue exploration into geometric reconstruction at high resolutions, refine the geometric performance of generated 3D models, and further enhance our framework.

\bibliographystyle{IEEEbib}
\bibliography{strings,refs}

\end{document}